\documentclass{article}

\usepackage[preprint]{neurips_2024}

\usepackage[utf8]{inputenc} 
\usepackage[T1]{fontenc}    
\usepackage{url}            
\usepackage{booktabs}       
\usepackage{amsfonts}       
\usepackage{nicefrac}       
\usepackage{microtype}      
\usepackage{xcolor}         
\usepackage{amsmath}
\usepackage{threeparttable}
\usepackage[pagebackref,breaklinks,colorlinks]{hyperref}
\usepackage{color,xcolor}
\usepackage{graphicx}
\usepackage{bbm}
\usepackage[capitalize]{cleveref}
\colorlet{dark-blue}{blue!70!black}
\hypersetup{
    colorlinks=true,%
    citecolor=dark-blue,%
    filecolor=dark-blue,%
    linkcolor=red,%
    urlcolor=magenta
}
\usepackage{color}
\usepackage{colortbl}
\definecolor{mygray}{gray}{.9}
\usepackage{bbding}
\usepackage{bm}
\usepackage{subcaption}

\usepackage{floatrow}
\newfloatcommand{capbtabbox}{table}[][\FBwidth]
\usepackage{multirow}
\usepackage{float}

\newfloatcommand{figurebox}{figure}[\nocapbeside][\dimexpr(\textwidth-\columnsep)/2\relax]

\newfloatcommand{tablebox}{table}[][\dimexpr(\textwidth-\columnsep)/2\relax]

\title{FreeTumor: Advance Tumor Segmentation via Large-Scale Tumor Synthesis}
\author{%
  Linshan Wu~~~~~Jiaxin Zhuang~~~~~Xuefeng Ni~~~~~Hao Chen\thanks{Corresponding author}\vspace{1mm}\\
  The Hong Kong University of Science and Technology\\
}
\begin{document}

\maketitle

\begin{abstract}
AI-driven tumor analysis has garnered increasing attention in healthcare. However, its progress is significantly hindered by the lack of annotated tumor cases, which requires radiologists to invest a lot of effort in collecting and annotation. In this paper, we introduce a highly practical solution for robust tumor synthesis and segmentation, termed~\emph{\textbf{FreeTumor}}, which refers to annotation-free synthetic tumors and our desire to free patients that suffering from tumors. Instead of pursuing sophisticated technical synthesis modules, we aim to design a simple yet effective tumor synthesis paradigm to unleash the power of large-scale data. Specifically, FreeTumor advances existing methods mainly from three aspects: \textbf{(1)} Existing methods only leverage small-scale labeled data for synthesis training, which limits their ability to generalize well on unseen data from different sources. To this end, we introduce the adversarial training strategy to leverage large-scale and diversified unlabeled data in synthesis training, significantly improving tumor synthesis. \textbf{(2)} Existing methods largely ignored the negative impact of low-quality synthetic tumors in segmentation training. Thus, we employ an adversarial-based discriminator to automatically filter out the low-quality synthetic tumors, which effectively alleviates their negative impact. \textbf{(3)} Existing methods only used hundreds of cases in tumor segmentation. In FreeTumor, we investigate the data scaling law in tumor segmentation by scaling up the dataset to \textbf{\emph{11k}} cases. Extensive experiments demonstrate the superiority of FreeTumor, \emph{e.g.}, on three tumor segmentation benchmarks, average $+8.9\%$ DSC over the baseline that only using real tumors and $+6.6\%$ DSC over the state-of-the-art tumor synthesis method. Code will be available.

\end{abstract}

\section{Introduction}
\label{sec_intro}

Tumor segmentation is one of the most fundamental tasks in medical image analysis~\cite{MSD,abdomenct1k,H-Dense,ZePT,mmwhs}, which has received significant attention recently. However, existing methods~\cite{H-Dense,swin,nnunet,unetr,VoCo} heavily rely on well-annotated tumor cases for training, which is time-consuming and requires extensive medical expertise~\cite{wang2021development,zhou2021review}. Thus, suffering from the annotation burden, the limited scale of tumor datasets significantly impedes the development of tumor segmentation.

To address the dilemma, tumor synthesis has become a burgeoning research topic recently~\cite{Syntumor,Difftumor,pixel2cancer}. Early attempts~\cite{lyu2022pseudo,yao2021label,han2019synthesizing,jin2021free,wang2022anomaly,wyatt2022anoddpm} utilized handcrafted image processing techniques to synthesize tumors. However, these synthetic tumors still differ significantly from real tumors~\cite{Syntumor}, and thus fail to improve segmentation performance effectively. SynTumor~\cite{Syntumor} liver tumor characteristics and designed several operations for tumor synthesis. However, handcrafting these characteristics requires significant expertise and is also limited to specific tumor types~\cite{Difftumor}. DiffTumor~\cite{Difftumor} proposed to train a conditioned diffusion model~\cite{ldm,vqgan} to synthesize tumors by reconstructing from Gaussian noises. Although with promising results, this conditioned diffusion model can only be trained when annotated tumors are available, thus the synthesis training is still limited by the scale of datasets and fails to generalize well on large-scale unseen datasets from other sources. More importantly, the synthetic results are not always perfect. However, the previous methods~\cite{Syntumor,pixel2cancer,Difftumor} have largely overlooked the negative impact of low-quality synthetic tumors on segmentation training.



\begin{figure}
	\centering
	\includegraphics[width=1\linewidth]{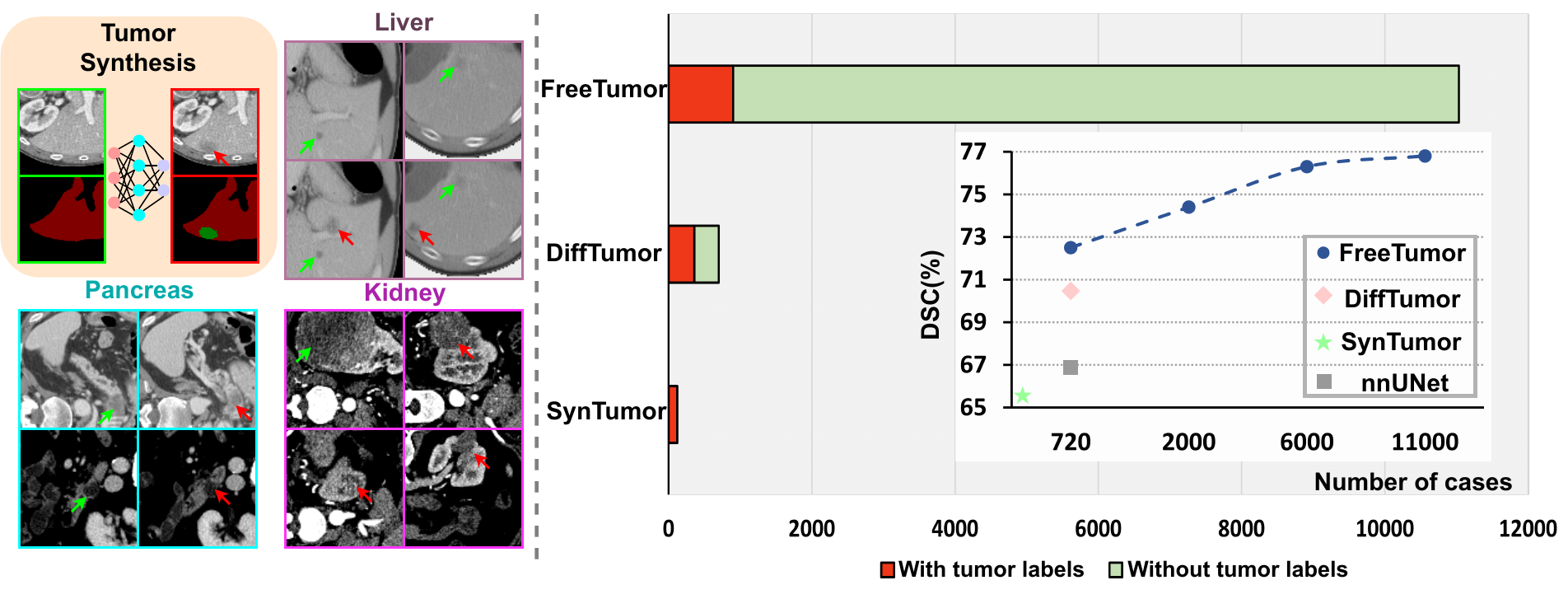}
	\caption{Synthesize different types of tumors with AI. The \textcolor{green}{green} and \textcolor{red}{red} arrows point to the real and synthetic tumors, respectively. We collect 0.9k labeled and 10k unlabeled data to facilitate tumor segmentation. By unleashing the power of large-scale data, FreeTumor outperforms the previous methods DiffTumor~\cite{Difftumor}, SynTumor~\cite{Syntumor}, and nnUNet~\cite{nnunet} by a significant margin.
 \vspace{-.1in}
 }
	\label{fig_intro}
\vspace{-.15in}
\end{figure}


In this work, we aim to \emph{unleash the power of large-scale data} to construct a stronger tumor segmentation model. The key challenge is to synthesize high-quality tumors on large-scale unlabeled data (Here, unlabeled represents without tumor labels as in~\cite{Syntumor,Difftumor}). Although the research of generative models, especially the Diffusion models~\cite{ldm,vqgan,ddpm,controlnet} have achieved astonishing conditioned image synthesis ability in the field of natural images, when applied to tumor synthesis, it is difficult to collect adequate annotated tumor cases to train a diffusion model~\cite{Difftumor}. Thus, limited by the data scale, it may fail to benefit large-scale unseen datasets with different characteristics, \emph{e.g.}, intensity, spacing, and resolution. In addition, to serve the following tumor segmentation, it is important to further design an automatic engine to discard low-quality synthetic tumors.

Thus, the off-the-shelf GAN-based image synthesis methods~\cite{SPADE,oasis,GAN,pix2pix,cyclegan} came into our view. GAN-based synthesis can effectively leverage \textbf{\emph{unpaired data}} for synthesis, which leads to a feasible way for us to leverage large-scale unlabeled data in tumor synthesis. There are two appealing advantages of GAN-based generative models in tumor synthesis: \textbf{(1)} The adversarial training strategy enables us to involve large-scale unlabeled data into synthesis training, \emph{i.e.}, generate tumors on unlabeled images and discriminate them with a discriminator (real or fake tumors). \textbf{(2)} The adversarial training strategy natively enables us to filter out low-quality synthetic results with the discriminator~\cite{oasis,freemask}, \emph{i.e.}, synthetic tumors failing to pass the discriminator will be discarded, thus significantly alleviating their negative impact in segmentation training. 


To this end, we present a highly practical solution for tumor synthesis and segmentation, termed \textbf{\emph{FreeTumor}}, which unleash the power of large-scale unlabeled data by high-quality tumor synthesis. Specifically, FreeTumor consists of three stages: \textbf{(1)} Training a baseline tumor segmentation model on labeled data, which serves as the discriminator in the following generative model. \textbf{(2)} Training a GAN-based tumor synthesis model, which simultaneously leverages labeled and large-scale unlabeled data in synthesis training and automatically discards low-quality synthetic tumors by the segmentation-based discriminator. \textbf{(3)} Training a tumor segmentation model with labeled and unlabeled data simultaneously, where the artificial tumors are synthesized in an online manner during training, as shown in Fig.~\ref{fig_intro}(a). Despite its simplicity, FreeTumor effectively leverages the large-scale unlabeled data for tumor segmentation training, significantly improving tumor segmentation performance.

Equipped with FreeTumor, we explore tumor segmentation with large-scale data. We collect 0.9k labeled and 10k unlabeled data to build an \textbf{11k} dataset for tumor segmentation, as shown in Fig.~\ref{fig_intro}(b). To the best of our knowledge, this is the \textbf{existing largest dataset} adopted for tumor synthesis and segmentation. We investigate the data scaling law by gradually scaling up the dataset and the results demonstrate the effectiveness. Extensive experimental results on three tumor segmentation benchmarks, \emph{i.e.}, LiTs~\cite{lits}, MSD-Pancreas~\cite{MSD}, and KiTs~\cite{kits}, demonstrate the effectiveness of our proposed FreeTumor. Specifically, compared with the baseline that used only real tumors for training, FreeTumor brings an average $8.9\%$ DSC improvements. FreeTumor also surpasses the state-of-the-art tumor synthesis method DiffTumor~\cite{Difftumor} by $6.6\%$ DSC. We also evaluate the results on the FLARE23 tumor segmentation public leaderboard and outperform existing methods~\cite{Syntumor,Difftumor} by a large margin. The 11k synthetic tumor dataset will be released to facilitate the research of tumor segmentation.


\section{Related Work}
\label{sec_Related_Work}
\vspace{-.01in}

\textbf{Generative models}. Generative models~\cite{GAN,ldm,ddpm} have witnessed rapid development in recent years, which can be roughly divided into two categories, \emph{i.e.}, unconditional~\cite{ho2022video,ho2022classifier} and conditional~\cite{controlnet,Difftumor,freemask}. Specifically, the unconditional ones only need input noise, while the conditional models can be controlled by text or labels and require these conditional inputs consistently. Among the structures of generative models, Generative Adversarial Networks (GANs)~\cite{GAN,pix2pix,cyclegan}, Variational AutoEncoder (VAE)~\cite{vqvae}, and diffusion models~\cite{ddpm,ldm,controlnet} have shown astonishing ability in synthesizing photo-realistic images. These generative models are also effectively applied to medical images~\cite{kazerouni2023diffusion,wu2024medsegdiff,dar2019image,chen2019synergistic,chen2020unsupervised,wolleb2022diffusion}, \emph{i.e.}, image-to-image
translation~\cite{li2023zero, ozbey2023unsupervised}, image reconstruction~\cite{assemlal2011recent,ma2021structure}, image denoising~\cite{chung2022mr, geng2021content}, and anomaly detection~\cite{siddiquee2019learning,Xiang_2023_CVPR}.

Although conditioned diffusion models have attracted more attention recently, they may require large-scale paired data for training~\cite{ldm,ddpm}, which is not very feasible in medical images. Compared to complex and diversified natural image synthesis, tumor synthesis simply requires low-level abnormal noise generation~\cite{Syntumor,Difftumor,pixel2cancer,yao2021label}. Thus, the traditional adversarial-based GAN methods can also fulfill the requirement. More importantly, \textbf{\emph{the usage of unpaired data}} in GAN leads to a potential way for us to leverage large-scale unlabeled data, and \textbf{\emph{the adversarial training strategy}} can natively assist in filtering the quality of synthetic images. Thus, in this paper, we explore the GAN-based generative models in tumor synthesis. Instead of pursuing sophisticated modules to generative models, we aim to develop a novel tumor synthesis paradigm to unleash the power of unlabeled data.

\textbf{Tumor synthesis}. Tumor synthesis has become an attractive topic in various medical modalities~\cite{Difftumor}, \emph{e.g.}, colonoscopy videos~\cite{shin2018abnormal}, Magnetic Resonance Imaging (MRI)~\cite{billot2023synthseg}, Computed Tomography (CT)~\cite{han2019synthesizing,lyu2022pseudo,yao2021label}, and endoscopic images~\cite{du2023boosting}. Early attempts~\cite{lyu2022pseudo, yao2021label,han2019synthesizing,jin2021free,wang2022anomaly,wyatt2022anoddpm} explored the low-level image processing techniques to synthesize tumors. However, there is still a large visual margin between the resulting synthetic tumors and the real tumors~\cite{Syntumor}. These synthetic tumors still cannot boost the tumor segmentation effectively, since the noisy generations will significantly deteriorate the model training. Thus, more advanced synthesis techniques are required.

SynTumor~\cite{Syntumor} sought to address this challenge by introducing a series of handcrafted operations, \emph{i.e.}, ellipse generation, elastic deformation, salt-noise generation, and Gaussian filtering. However, it requires extensive expertise of radiologists to handcraft the characteristics, while it is still restricted to a specific type of tumor (liver tumor in the paper~\cite{Syntumor}) and fails to generalize to other types of tumors~\cite{Difftumor}. DiffTumor~\cite{Difftumor} further proposed to employ conditional diffusion models~\cite{ldm,vqgan} to synthesize tumors. Although promising results have been demonstrated, there are still two main problems: \textbf{(1)} The synthesis training of DiffTumor~\cite{Difftumor} is mainly based on reconstructing CT volumes with annotated tumors, which is still heavily restricted by the scale of annotated tumor dataset. Although DiffTumor~\cite{Difftumor} leveraged 10k data to train the autoencoder~\cite{vqgan}, in the synthesis training only $0.3k$ labeled tumor data are used. Thus, when adapting it to \textbf{\emph{large-scale unseen datasets}} from other sources, the synthesis performance will significantly decrease. \textbf{(2)} More importantly, although most of the synthetic tumors are realistic in comparison to real liver tumors, the synthesis results cannot be always perfect~\cite{pixel2cancer}. The failure cases will significantly deteriorate the subsequent tumor segmentation training~\cite{freemask}. Thus, to alleviate the negative impact of low-quality synthetic tumors, it is vital to further design a selection strategy to automatically filter out the unsatisfactory synthesis.

To this end, we present FreeTumor, a GAN-based tumor synthesis paradigm, with two main advantages upon previous methods: \textbf{(1)} A well-designed adversarial training strategy to leverage unpaired data, thus unleashing the power of large-scale unlabeled data for synthesis training. \textbf{(2)} An adversarial-based discriminator to filter out the unsatisfactory synthetic tumors, thus alleviating the negative impact of low-quality synthetic tumors in tumor segmentation.



\textbf{Learning from synthetic images}. Learning from synthetic images is a commonly-used technique in label-efficient learning~\cite{freemask,CISC_R,mim,AGMM,DBFNet,agmm++}, aiming to ease the burden of collecting large-scale dataset and annotating dense labels. The main sources of generating synthetic images can be roughly divided into three categories~\cite{freemask}, \emph{i.e.}, computer graphics engines~\cite{richter2016playing,ros2016synthia}, image processing techniques~\cite{zhong2020random,zhong2022adversarial,baradad2021learning,kataoka2022replacing,takashima2023visual,DCA}, and generative models~\cite{freemask,souly2017semi,zhang2021datasetgan,sariyildiz2023fake,wang2024do}. After synthesis, collaborative training~\cite{freemask,depthanything,yang2022st++,yang2023revisiting,yang2023shrinking,liu2023multi} is used to combine the real and synthetic images. In this paper, we aim to explore this paradigm for tumor segmentation training.

\section{Method}
\label{sec_method}

\begin{figure}
	\centering
	\includegraphics[width=1\linewidth]{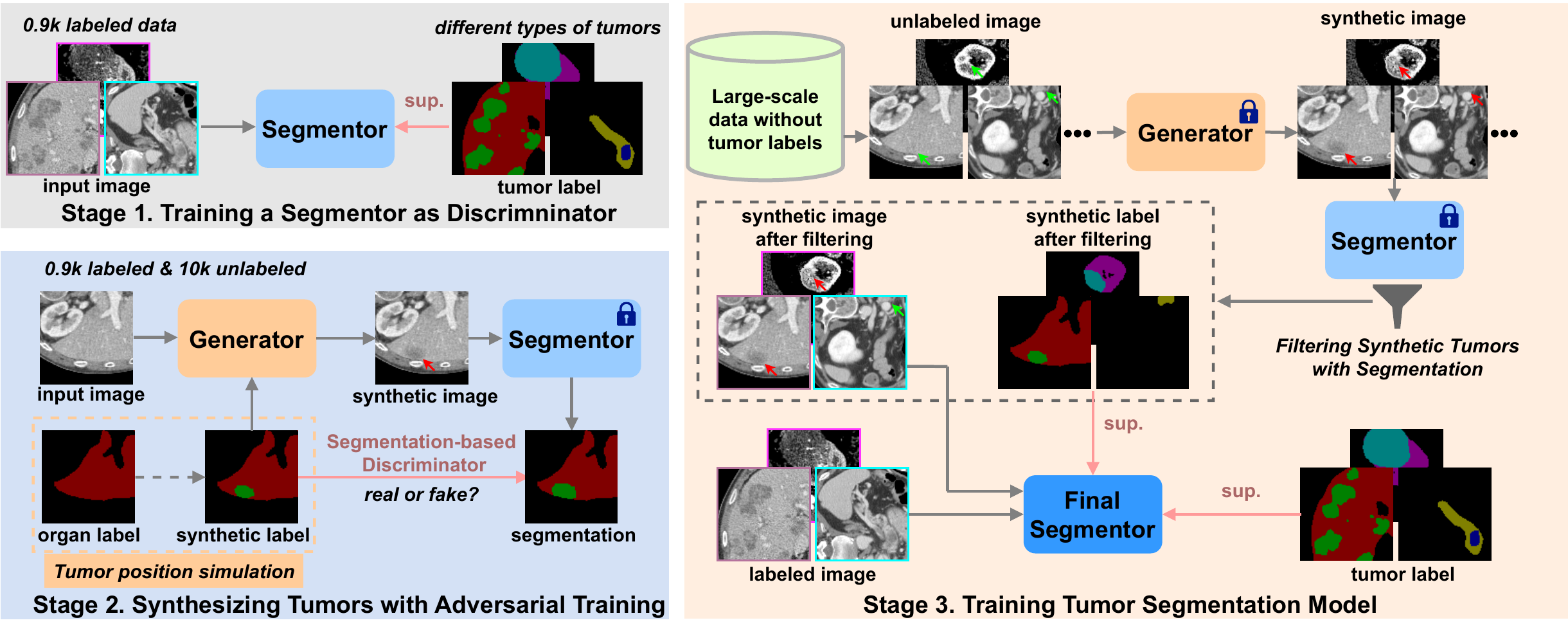}
	\caption{\textbf{The overall framework of FreeTumor}, including three stages: \textbf{(1)} \emph{Training a segmentor as discriminator}. We first leverage the labeled data to train a baseline segmentor as the discriminator in the following generative model. \emph{Sup.} denotes supervision. \textbf{(2)} \emph{Synthesizing tumors with adversarial training}. The tumor position simulation follows~\cite{Syntumor,Difftumor}. We leverage both labeled and unlabeled data to train a tumor synthesis model. Thanks to the previous efforts~\cite{abdomenatlas,abdomenct1k,MSD,FLARE22} in collecting data, we can formulate an \textbf{11k} dataset with organ labels for tumor position simulation. Motivated by~\cite{oasis}, we use the baseline segmentor to discriminate the reality of synthetic tumors. \textbf{(3)} \emph{Training tumor segmentation model}. We generate and filter the synthetic tumors on the large-scale data for training the final tumor segmentor.}
	\label{fig_frame}
\vspace{-.1in}
\end{figure}

In this section, we first introduce our tumor synthesis pipeline in Section~\ref{sec_method_synthesis}. Then, we further clarify our strategy to filter out low-quality synthetic tumors in Section~\ref{sec_method_filter}. Following this, we present the process of unleashing the power of large-scale unlabeled data with tumor synthesis in Section~\ref{sec_method_unleash}. Lastly, in Section~\ref{sec_method_foundation}, we further discuss the paradigms to leverage synthetic tumors for tumor segmentation training. The overall framework of \emph{FreeTumor} is shown in Fig.~\ref{fig_frame}.

\subsection{Synthesizing Tumors with Adversarial Training}
\label{sec_method_synthesis}

Motivated by OASIS~\cite{oasis}, we propose to leverage a baseline tumor segmentation model to discriminate real and fake tumors, as shown in Fig.~\ref{fig_frame}. In Stage 1, we first train a baseline segmentor with only labeled data (with annotated tumors), which will be employed as a discriminator in the following generative model to discriminate the synthetic tumors.

In Stage 2, we employ the adversarial training strategy to train a tumor generative model. The first step is to simulate the tumor positions on the organs, which aims to select a proper location for the synthetic tumors. For fair comparisons, we follow the pipeline of SynTumor~\cite{Syntumor} and DiffTumor~\cite{Difftumor} in tumor position simulation. Since the organ labels already exist in the collected dataset, it is easy to select a location to synthesize tumors. Here, we denote the tumor mask as $M$ that represents the positions of synthetic tumors, where $M=1$ are the positions of synthetic tumors and $M=0$ are remained as the original values.

The Generator $G$ used in Stage 2 is a typical encoder-decoder based U-Net~\cite{UNET}, which is widely used in current state-of-the-art generative models~\cite{freestyle,oasis,Difftumor,ldm,freemask}. In FreeTumor, we aim to use the Generator $G$ to transform the voxel values from organ to tumor. Specifically, we use $x$ to denote the original image, $\hat{x}$ denotes the synthetic image, the transform process is as follows:
\begin{equation}\label{eq_synthetic}
    \hat{x} = (1-M){\otimes}x + M{\otimes}[x - tanh(G(x)){\otimes}g(x)],
\end{equation}
where $x$ is first normalized to $0{\sim}1$ and $g(x)$ is the Gaussian filter to blur the texture~\cite{Syntumor}. $tanh$ is the activation function to normalize $G(x)$. It can be seen that with the tumor mask $M$, only the synthetic positions are transformed and other positions are reserved as the original values.

According to Eq.(\ref{eq_synthetic}), FreeTumor synthesizes tumors by estimating the distance ($tanh(G(x))$) between organs and tumors, which is inherited from the observation of the previous work SynTumor~\cite{Syntumor}. However, SynTumor~\cite{Syntumor} empirically set a fixed value as the distance, which heavily limited the ability of generalization. Thus, we make it a learnable process in FreeTumor, which is more robust. DiffTumor~\cite{Difftumor} synthesized tumors by directly reconstructing values of tumors while ignoring to consider the information of corresponding organs and backgrounds, thus may fail to generalize well to unseen datasets with different value intensities.

In FreeTumor, we propose to employ a tumor segmentation model for adversarial training. Specifically, we first freeze the parameters of the baseline segmentor $S$ trained in Stage 1, then feed the synthetic images to the frozen segmentor $S$. We aim to use the segmentation results of these synthetic images to optimize the Generator by adversarial training. Concretely, it is intuitive that if a case of synthetic tumor appears realistic in comparison to the real tumors, it has a higher probability to be segmented by the baseline segmentor $S$. This observation is also explored by OASIS~\cite{oasis}. Motivated by this, we calculate the segmentation loss $L_{seg}$ for adversarial training as follows:
\begin{equation}\label{loss_seg}
    L_{seg} = \frac{1}{\parallel M\parallel}\sum_{M=1}{\parallel 1 - S(\hat{x})\parallel},
\end{equation}
where $S(\hat{x})$ is the tumor prediction logits generated by the baseline segmentor $S$, and we employ the simplest Euclidean distance to optimize the generator $G$. 

In addition, following the traditional GAN~\cite{GAN,cyclegan,SPADE,oasis,pix2pix}, besides the segmentor, we also adopt another classifier discriminator $C$ to discriminate real or fake tumors using a typical classification loss $L_{cls}$. The classifier $C$ works similarly to the previous adversarial training~\cite{GAN,cyclegan,SPADE,oasis,pix2pix}: (1) In the discriminating process, $C$ is optimized to distinguish real and synthetic tumors. (2) In the generating process, $C$ is frozen and tries to classify the synthetic tumors as the real tumors, thus optimizing the generator $G$. Thus, the total adversarial training loss $L_{adv}$ is as follow:
\begin{equation}\label{loss_adv}
    L_{adv} = \mathop{max}_{G^{\sim}}\mathop{min}_{D^{\sim}}\lambda_{cls}L_{cls} + {L_{seg}},
\end{equation}
where $G^{\sim}$ and $D^{\sim}$ represent the generating and discriminating processes, respectively. $\lambda_{cls}$ is the weight of $L_{cls}$ and is set to $0.1$ in experiments empirically.

\textbf{Replace with diffusion model}. The generator can be replaced by a diffusion model~\cite{adversarialdiffusion,nie2022diffusion}. The difference is that we use adversarial training to filter the synthetic tumors generated by the diffusion model. We explore the filtering strategy to select the results of the trained DiffuTumor~\cite{Difftumor}. Empirically, we observe that when replacing DiffTumor~\cite{Difftumor} as the generator, the performances did not increase, which suggests that the usage of large-scale data and the filtering strategy are more important. 


\subsection{Filtering Synthetic Tumors with Segmentation}
\label{sec_method_filter}
\textbf{Motivation}. Although promising synthesis performances have been demonstrated~\cite{Syntumor,Difftumor}, the synthetic tumor cannot always be perfect. We observe that the low-quality synthetic tumors will significantly deteriorate the tumor segmentation results. Both SynTumor~\cite{Syntumor} and DiffTumor~\cite{Difftumor} ignored to alleviate these negative impacts. Although they further invite experienced radiologists to review the synthesis results, it still requires laborious efforts and cannot be applied to large-scale data. Thus, we develop an effective filtering strategy to automatically discard low-quality synthetic tumors.

\textbf{Segmentation-based discriminator for filtering}. Our filtering strategy is based on the segmentation-based discriminator, which is also one of the strong reasons why we adopted the adversarial training strategy instead of diffusion models to synthesize tumors. Inspired by FreeMask~\cite{freemask}, we propose to adaptively filter low-quality synthetic tumors by calculating the \textbf{\emph{proportions of satisfactory synthesized tumor regions}}. The satisfactory synthesized tumors represent the synthetic tumors that do match the corresponding tumor masks $M$ well. Intuitively, we can use the baseline segmentor $S$ to calculate the correspondence: the proportions of synthetic tumors that are segmented as tumors. Thus, we calculate the proportion $P$ as follows:
\begin{equation}\label{eqn_proportion}
    P = \frac{\sum_{i=1}^{N}[\mathbbm{1}(S({\hat{x}})){\times}\mathbbm{1}(M=1))]}{\sum_{i=1}^{N}[\mathbbm{1}(M=1))]},
\end{equation}
where $N$ denotes the total number of voxels, $\mathbbm{1}(S({\hat{x}})$ denotes the number of voxels that are segmented as tumors, $\mathbbm{1}(M=1))$ denotes the number of voxels that tumor mask is $1$ (the positions of synthetic tumors). It is intuitive that \emph{if the proportion} $P$ \emph{is higher, the quality of this case of synthetic tumor tends to be higher}~\cite{oasis}. Although this measurement cannot be absolutely accurate, it still largely reveals the tumor synthesis results, which can serve as an automatic tool for quality test.

\begin{figure}
	\centering
	\includegraphics[width=1\linewidth]{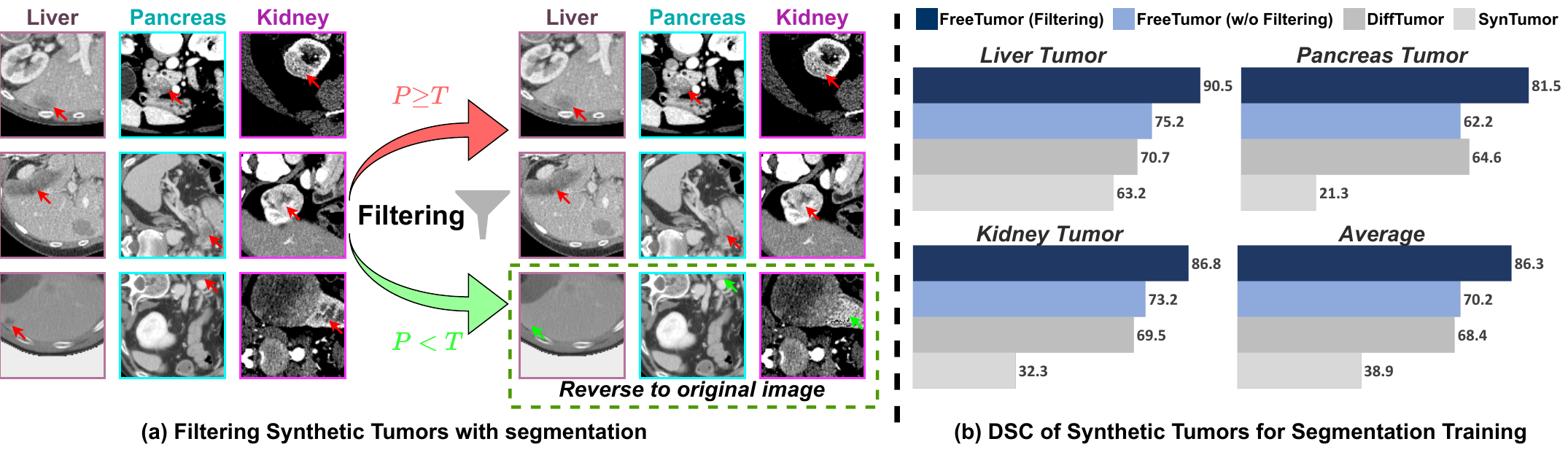}
	\caption{\textbf{Segmentation-based Filtering strategy for synthetic tumors}. (a) We discard the unsatisfactory synthetic tumors according to Eq.~(\ref{eqn_turing}). (b) We use the baseline segmentor $S$ to test the accuracy of synthetic tumors, verifying with the segmentation DSC. It can be seen that with our proposed filtering strategy, the DSC of synthetic tumors are improved significantly (average $+16.1\%$), which also surpass SynTumor~\cite{Syntumor} and DiffTumor~\cite{Difftumor} by a large margin.}
	\label{fig_filtering}
\vspace{-.1in}
\end{figure}

We set a threshold $T$ to split the high- and low-quality synthetic tumors. We use the term \textbf{\emph{Quality Test}} to represent whether the synthetic case passes the discriminator, the Filtering strategy $F$ is defined as:
\begin{equation}\label{eqn_turing}
F(x|P, T, G, S)=\left\{
     \begin{array}{lr}
     \hat{x}, ~~P{\geq}T,  &\text{This synthetic tumor pass the Quality Test}\\
     x, ~~P{<}T,  &\text{This synthetic tumor fail the Quality Test}
     \end{array}
\right.
\end{equation}
With $F$, we effectively filter the synthetic tumors online, as shown in Fig.~\ref{fig_filtering}. Despite its simplicity, we effectively alleviate the negative impact of unsatisfactory synthetic tumors in segmentation training, which is a significant improvement upon the previous methods~\cite{Syntumor,Difftumor}. 


\subsection{Unleashing the Power of Unlabeled Data with Tumor Synthesis}
\label{sec_method_unleash}

Distinguished from previous works that use a limited scale of dataset for tumor segmentation training, we emphasize the importance of unlabeled images in the development of tumor segmentation. Thanks to prior efforts~\cite{abdomenatlas,abdomenct1k,li2023well,FLARE22} in building large-scale CT datasets, we can easily collect adequate CT images for training. The challenge is that these datasets only contain organ labels while lacking annoated tumor cases. Thus, we develop FreeTumor to unleash the power of these unlabeled data.

In this paper, `labeled' represents `with tumor labels' and `unlabeled' represents `without tumor labels' but the organ labels are available, which follows the previous works~\cite{Syntumor,Difftumor,pix2pix}. With the organ labels, we can easily generate and filter synthetic tumors as described in Secs~\ref{sec_method_synthesis} and \ref{sec_method_filter}. Specifically, given the unlabeled dataset $D_{u}$, we conduct tumor synthesis for $D_{u}^{'}$ as follow:
\begin{equation}\label{eqn_syn_unlabeled}
    D_{u}^{'} = \left\{(x, F[G(x)], S) | x{\in}{D_{u}}\}\right..
\end{equation}

\textbf{Online tumor synthesis}. Unlike the previous method~\cite{freemask}, we synthesize tumors online during training as~\cite{Syntumor,Difftumor}. This is because online generation allows us to synthesize more diversified tumor cases, thus improving the robustness of tumor segmentation. We have further compared the effectiveness of online and offline generation in the experiments (see appendix). We will also release our offline 11k synthetic tumor dataset to facilitate the following research. Note that \emph{\textbf{tumor synthesis is also conducted on the labeled data}}, since the tumors in the labeled data are also rare~\cite{Syntumor}. 


\textbf{Mixup for unlabeled data}. To collect a large-scale dataset, there inevitably exists domain gaps between different sources of datasets, \emph{e.g.}, intensity, spacing, and resolution. Motivated by previous works~\cite{cutmix,depthanything}, we propose to mix up~\cite{cutmix} different sources of datasets as follows:
\begin{equation}\label{eqn_mixup}
    x_{ab} = {\hat{x}}_{a}{\otimes}m+{\hat{x}}_{b}{\otimes}(1-m), {\hat{x}}_{a}, {\hat{x}}_{b}{\in}D_{u}^{'},
\end{equation}
where $m$ is the mask for Mixup. The effectiveness of Mixup are discussed in the experiments.

\subsection{Leveraging Synthetic Tumors for Tumor Segmentation Training}
\label{sec_method_foundation}
\textbf{Pretrain-Finetune vs Jointly-Training}. We further explore the training paradigm for combining the labeled and unlabeled data in tumor segmentation training. Following the previous works~\cite{freemask,depthanything,Difftumor}, we discuss two training strategies: \textbf{(1)} Pre-training on large-scale synthetic tumor datasets then finetune on the labeled dataset. \textbf{(2)} Jointly training labeled and synthetic data simultaneously. Since the unlabeled dataset is of a much larger scale, we over-sample the labeled data to the same scale. We found that jointly training achieves better performance empirically. 

\textbf{Universal vs Specialist models}. Previous works~\cite{nnunet,Syntumor,Difftumor,pix2pix} proposed to train multiple specialist models for different types of tumors, \emph{i.e.}, liver tumor model for liver tumors and pancreas tumor model for pancreas tumor model, respectively. However, it is more efficient to use one universal model to solve different types of tumors simultaneously. Thus, we further compare the universal model with specialist models~\cite{nnunet,Syntumor,Difftumor,pix2pix}. We observe that with small-scale datasets ($<2k$) for training, ensembling multiple specialist models together can surpass one universal model. However, when we scale up the dataset, one universal model can achieve competitive performances with multiple specialist models, while requiring no multiple times of training for different types of tumors. FreeTumor also demonstrates a strong \textbf{zero-shot transfer ability} on different types of tumors, \emph{i.e.}, train with one specific type of tumor then transfer to other types of tumors.

\section{Experiments}
\label{sec_experi}

\begin{table}
	\setlength{\abovecaptionskip}{0.pt}
	\setlength{\belowcaptionskip}{-0.em}
	\centering
	\footnotesize
 \caption{Overall performances on three tumor segmentation benchmarks~\cite{lits,MSD,kits}. We report the tumor segmentation DSC(\%). $\dagger$ denotes we re-implement the method with the same settings as~\cite{Difftumor}, other results are directly drawn from the original papers~\cite{Difftumor,Syntumor,pixel2cancer}. Extra data and post-processing are considered. The results of baseline SwinUNETR~\cite{swin} and the SOTA DiffTumor~\cite{Difftumor} are \underline{underlined}. The best results are \textbf{bolded}. 720~\cite{Difftumor} is a cherry-picked sub-set of 0.9k (see appendix).}
 \vspace{-.1in}
\begin{threeparttable}
	\begin{tabular}{c|cc|ccc|c}
		\toprule[1.2pt]
		\textbf{Method} &\textbf{Ext. Data} &\textbf{Post-Pro.} &\textbf{Liver} &\textbf{Panc.} &\textbf{Kidney} &\textbf{Average}\\
		\hline
        \textbf{\emph{Without Synthetic Tumors}} & & & & & &\\
        U-Net~\cite{UNET} &\XSolidBrush &\XSolidBrush &62.5 &51.2 &72.0 &61.9\\
        nnUNet~\cite{nnunet} &\XSolidBrush &\CheckmarkBold &62.9 &53.7 &76.9 &64.5\\
        UNETR~\cite{unetr} &\XSolidBrush &\XSolidBrush &60.3 &50.6 &74.1 &61.7\\
        SwinUNETR~\cite{swin} &\XSolidBrush &\XSolidBrush &61.8 &52.9 &74.6 &63.1\\
        SwinUNETR$\dagger$~\cite{swin} &\XSolidBrush &\XSolidBrush &\underline{64.7} &\underline{59.0} &\underline{80.3} &\underline{67.9}\\
        VoCo~\cite{VoCo} &\CheckmarkBold &\XSolidBrush &65.5 &59.9 &80.8 &68.7\\
        \hline
        \textbf{\emph{With Synthetic Tumors}} & & & & & &\\
        CopyPaste$\dagger$~\cite{copypaste} &\XSolidBrush &\XSolidBrush &60.2 &39.8 &65.2 &55.1\\
        Yao~\emph{et al.}~\cite{yao2021label} &\XSolidBrush &\XSolidBrush &32.8 &- &- &-\\
        SynTumor~\cite{Syntumor} &\XSolidBrush &\XSolidBrush &59.8 &- &- &-\\
        SynTumor$\dagger$~\cite{Syntumor} &\XSolidBrush &\XSolidBrush &66.2 &42.5 &67.9 &58.9\\
        Pixel2Cancer~\cite{pixel2cancer} ($\sim$1k) &\CheckmarkBold &\XSolidBrush &56.7 &59.3 &73.9 &63.3\\
        DiffTumor~\cite{Difftumor} (720) &\CheckmarkBold &\CheckmarkBold &\underline{67.9} &\underline{61.0} &\underline{81.8} &\underline{70.2}\\
        \rowcolor{mygray}
        FreeTumor (0.9k) &\XSolidBrush &\XSolidBrush &71.2 &63.1 &84.3 &72.9\\
        \rowcolor{mygray}
        FreeTumor (2.2k) &\CheckmarkBold &\XSolidBrush &72.5 &65.2 &85.6 &74.4\\
        \rowcolor{mygray}
        FreeTumor (6k) &\CheckmarkBold &\XSolidBrush &74.3 &67.4 &86.7 &76.3\\
        \rowcolor{pink}
        FreeTumor (11k) &\CheckmarkBold &\XSolidBrush &\textbf{74.5} &\textbf{68.6} &\textbf{87.3} &\textbf{76.8}\\
        \hline
        \textcolor{cyan}{\bm{$\triangle$}(SwinUNETR)} & & &\textcolor{cyan}{\bm{$\uparrow$$9.8\%$}} &\textcolor{cyan}{\bm{$\uparrow$$9.6\%$}} &\textcolor{cyan}{\bm{$\uparrow$$7.0\%$}} &\textcolor{cyan}{\bm{$\uparrow$$8.9\%$}}\\
        
        \textcolor{cyan}{\bm{$\triangle$}(DiffTumor)}& & &\textcolor{cyan}{\bm{$\uparrow$$6.6\%$}} &\textcolor{cyan}{\bm{$\uparrow$$7.6\%$}} &\textcolor{cyan}{\bm{$\uparrow$$5.5\%$}} &\textcolor{cyan}{\bm{$\uparrow$$6.6\%$}}\\
        \toprule[1.2pt]
	\end{tabular}
    \end{threeparttable}        
\label{table_overall_performance}
\vspace{-.2in}
\end{table}

\subsection{Dataset}
\label{sec_exp_dataset}
The real tumor datasets are from LiTS~\cite{lits}, MSD-Pancreas~\cite{MSD}, and KiTs~\cite{kits}, which contain liver, pancreas, and kidney tumors, respectively. These three datasets are also used for validation. The datasets with only organ labels are from CHAOS~\cite{chaos}, Pancreas-CT~\cite{panc_ct}, BTCV~\cite{btcv}, Amos22~\cite{amos}, WORD~\cite{word}, Flare22~\cite{FLARE22}, Abdomenct-1k~\cite{abdomenct1k}, AbdomenAtlas~\cite{abdomenatlas}, and Flare23~\cite{FLARE22}. We adopt consistent experiment settings with DiffTumor~\cite{Difftumor} for fair comparisons. We further evaluate the results on the FLARE23 public leaderboard, which includes liver, pancreas, kidney, stomach, and colon tumors. Datasets, implementation, and visualization results are further described in the appendix.


\subsection{Overall Performance}
\label{sec_exp_performace}

As shown in Table~\ref{table_overall_performance}. We compare our method with multiple baseline methods (without synthetic tumors) and the synthetic methods. Specifically, U-Net~\cite{UNET}, nnUNet~\cite{nnunet}, UNETR~\cite{unetr}, and SwinUNETR~\cite{swin} are trained only with real tumors. We further evaluate the effectiveness of self-training~\cite{VoCo} in tumor segmentation. For all of the synthetic methods~\cite{yao2021label,pixel2cancer,Syntumor,Difftumor}, we use the SwinUNETR~\cite{swin} as the backbone for fair comparisons. DiffTumor~\cite{Difftumor} further used the universal model~\cite{clipdriven} to post-process the tumor predictions. Since the original SynTumor~\cite{Syntumor} was conducted on the liver tumors~\cite{lits} only, we re-implement it on pancreas~\cite{MSD} and kidney tumors~\cite{kits}.


Compared with the baseline SwinUNETR~\cite{swin}, FreeTumor brings an average of \bm{$8.9\%$} DSC improvements, which also surpass the previous state-of-the-art DiffTumor~\cite{Difftumor} by \bm{$6.6\%$} DSC. Specifically, self-supervised learning~\cite{VoCo} cannot bring obvious improvements in tumor segmentation, which suggests that tumor synthesis on unlabeled images is more effective. SynTumor~\cite{Syntumor} gained improvements in liver tumor segmentation but deteriorated the segmentation of pancreas and kidney tumors. This is because SynTumor~\cite{Syntumor} is specifically designed for liver tumors while failing to generalize to other types of tumors~\cite{Difftumor}. \textbf{The results of the FLARE23 public leaderboard are in the appendix}.

\subsection{Ablation Studies}
\label{sec_exp_ablation}

\begin{table*}
    \setlength{\abovecaptionskip}{-.1in}
	\setlength{\belowcaptionskip}{-.2in}
    \begin{floatrow}
    \capbtabbox{
        \begin{tabular}{ccccc}
        \toprule[1.2pt]
		\textbf{Method} &\textbf{Data} &\textbf{Unlab.} &\textbf{Train.} &\bm{$P(\%)$}\\
        \hline
        SynTumor~\cite{Syntumor} &131 &\XSolidBrush &\XSolidBrush &43.7\\
        DiffTumor~\cite{Difftumor} &360 &\XSolidBrush &\CheckmarkBold &67.0\\
        \hline
        \multirow{3}{*}{FreeTumor} &0.9k &\XSolidBrush &\CheckmarkBold &79.5\\
        &6k &\CheckmarkBold &\CheckmarkBold &81.8\\
        &11k &\CheckmarkBold &\CheckmarkBold &82.2\\
        \toprule[1.2pt]
        \end{tabular}
    }{
     \caption{Evaluation of $P$ in synthesis quality. \emph{Unlab.} denotes whether to use unlabeled data in synthesis training and \emph{Train.} denotes whether the synthesis is trainable.
     \vspace{-.1in}}
     \label{table_abla_proportion}
    }
    
    \capbtabbox{
    \begin{tabular}{ccccc}
    \toprule[1.2pt]
		\multicolumn{2}{c}{\textbf{Loss}} &\multirow{2}{*}{\textbf{Filt.}} &\multirow{2}{*}{$\dagger$\textbf{Diff.}} &\multirow{2}{*}{\textbf{DSC}}\\
        \cline{1-2}
        \bm{$L_{seg}$} &\bm{$L_{cls}$} &\\
		\hline
        \CheckmarkBold &\XSolidBrush &\CheckmarkBold &\XSolidBrush &74.0\\
        \CheckmarkBold &\CheckmarkBold &\CheckmarkBold &\XSolidBrush &\cellcolor{pink}74.4\\
        \hline
        \XSolidBrush &\XSolidBrush &\XSolidBrush &\CheckmarkBold &70.5\\
        \XSolidBrush &\XSolidBrush &\CheckmarkBold &\CheckmarkBold &72.8\\
        \toprule[1.2pt]
    \end{tabular}
    }{
     \caption{Evaluation of adversarial training (2.2k). $\dagger$ denotes using the trained diffusion model~\cite{Difftumor} as the generator.
     \vspace{-.1in}}
     \label{table_abla_generation}
     \small
    }
    \end{floatrow}
\vspace{-.1in}
\end{table*}

\begin{table*}
    \setlength{\abovecaptionskip}{-.1in}
	\setlength{\belowcaptionskip}{-.2in}
    \begin{floatrow}
    \capbtabbox{
        \begin{tabular}{cccc}
        \toprule[1.2pt]
		\textbf{Method} &\textbf{Synthetic} &\textbf{Filtering} &\textbf{DSC}\\
        \hline
        SwinUNETR~\cite{swin} &\XSolidBrush &\XSolidBrush &67.9\\
        \hline
        SynTumor~\cite{Syntumor} &\CheckmarkBold &\XSolidBrush &58.9\\
        DiffTumor~\cite{Difftumor} &\CheckmarkBold &\XSolidBrush &70.2\\
        \hline
        \multirow{4}{*}{FreeTumor} &\CheckmarkBold &\XSolidBrush &72.3\\
        &\CheckmarkBold &\bm{$T=0.5$} &73.5\\
        &\CheckmarkBold &\bm{$T=0.7$} &\cellcolor{pink}74.4\\
        &\CheckmarkBold &\bm{$T=0.9$} &74.0\\
        \toprule[1.2pt]
        \end{tabular}
    }{
     \caption{Evaluation of the threshold $T$ in filtering (2.2k).
     \vspace{-.15in}}
     \label{table_abla_filtering}
    }
    
    \capbtabbox{
    \begin{tabular}{cccc}
    \toprule[1.2pt]
		\textbf{Method} &\textbf{Seg.} &\textbf{Syn.} &\textbf{DSC}\\
        \cline{2-4}
        \multirow{5}{*}{FreeTumor} 
        &0.9k &0.9k &72.9\\
        &2k &2k &74.4\\
        \cline{2-4}
        &11k &0.9k &73.8\\
        &11k &2k &74.2\\
        &11k &11k &\cellcolor{pink}76.8\\
        \toprule[1.2pt]
        &\textbf{Seg.} &\textbf{Mixup} &\textbf{DSC}\\
        \cline{2-4}
        \multirow{4}{*}{FreeTumor} 
        &0.9k &\XSolidBrush &72.7\\
        &0.9k &\CheckmarkBold &\cellcolor{pink}72.9\\
        \cline{2-4}
        &11k &\XSolidBrush &73.3\\
        &11k &\CheckmarkBold &\cellcolor{pink}76.8\\
        \toprule[1.2pt]
    \end{tabular}
    }{
     \caption{Synthesis training scale and Mixup.
     \vspace{-.15in}}
     \label{table_abla_mixup}
     \small
    }
    
    \end{floatrow}
\vspace{-.2in}
\end{table*}

\textbf{Quality of synthetic tumors}. It is difficult to use image quality metrics to measure the tumor synthesis quality~\cite{Syntumor,Difftumor}. Previous works~\cite{Syntumor,Difftumor,pixel2cancer} hired radiologists to verify the synthetic results case by case, which is laborious. Motivated by OASIS~\cite{oasis}, we propose to use a tumor segmentation model~\cite{swin} to verify the quality of synthetic tumors as in Section~\ref{sec_method_filter}: proportions of synthetic tumors that are segmented by the segmentor. Although this strategy cannot be absolutely reliable, it can reveal the quality to some extent, as shown in Table~\ref{table_abla_proportion}. We observe that with more data for synthesis training, the performances are better. Note that DiffTumor~\cite{Difftumor} used 10k cases of CT to train the auto-encoder~\cite{vqvae}, but only 360 cases with tumor labels are used in synthesis training. 


\textbf{Adversarial training for generation}. In Table~\ref{table_abla_generation}, we evaluate $L_{seg}$ and $L_{cls}$ in synthesis. We replace the generator with a trained diffusion model~\cite{Difftumor} for comparison. We also use the filtering strategy to control the results of diffusion model~\cite{Difftumor}. It can be seen that the synthesis paradigm especially the filtering strategy is important.

\textbf{Effectiveness of filtering strategy}. In Table~\ref{table_abla_filtering}, we further evaluate the setting of threshold $T$ in the filtering strategy, as described in Section~\ref{sec_method_filter}. For efficiency, we use 2.2k data for this ablation study. It can be seen that the filtering strategy is important for improving the following tumor segmentation. Based on the results in Table~\ref{table_abla_filtering}, we set $T=0.7$ for all the experiments in FreeTumor.

\textbf{Synthesis training scale and Mixup}. As shown in Table~\ref{table_abla_mixup}, if we only leverage 11k data in segmentation training without using them in synthesis training, the improvements are marginal. Thus, it is important to scale up the data in both synthesis and segmentation training. Otherwise, we would synthesize tumors on unseen data in segmentation training and the performance will be hindered. Mixup also plays an important role when training with large-scale data. When scaling up to 11k data, the performance will drop without Mixup. We conclude that it is because 11k data covers different sources of datasets, and the domain gap among them may deteriorate the training. 



\begin{table*}
    \setlength{\abovecaptionskip}{-.1in}
	\setlength{\belowcaptionskip}{-.2in}
    \begin{floatrow}
    \capbtabbox{
        \begin{tabular}{ccc}
        \toprule[1.2pt]
		\textbf{Paradigm} &\textbf{Data} &\textbf{DSC}\\
        \hline
        Only Real Tumors  &0.9k &67.9\\
        Only Synthetic Tumors &10k &60.6\\
        \hline
        \textbf{\emph{Real and Synthetic}}\\
        Pretrain-Finetune &11k &72.7\\
        Jointly-Training &11k &\cellcolor{pink}76.8\\
        \toprule[1.2pt]
        \end{tabular}
    }{
     \caption{Pretrain-Finetune vs Jointly-Training.
     \vspace{-.15in}}
     \label{table_abla_training}
    }
    
    \capbtabbox{
    \begin{tabular}{cc|ccc|c}
    \toprule[1.2pt]
		\textbf{Method} &\textbf{Data} &\textbf{Liv.} &\textbf{Pan.} &\textbf{Kid.} &\textbf{AVG}\\
		\hline
        \multirow{2}{*}{Univers.}
        &0.9k &68.1  &60.9 &82.0 &70.3\\
        &11k &74.0  &68.2 &87.5 &76.6\\
        \hline
        \multirow{2}{*}{Special.}
        &0.9k &71.2 &63.1 &84.3 &72.9\\
        &11k &74.5 &68.6 &87.3 &\cellcolor{pink}76.8\\
        \toprule[1.2pt]
    \end{tabular}
    }{
     \caption{Universal vs Specialist models.
     \vspace{-.15in}}
     \label{table_abla_generalist}
     \small
    }
    
    \end{floatrow}
\vspace{-.15in}
\end{table*}

\begin{figure}
\begin{floatrow}
\CenterFloatBoxes
\ttabbox{%
  \centering
  \begin{tabular}{cccc}
    \toprule[1.2pt]
		\textbf{Method} &\textbf{Data} &\textbf{Synthetic} &\textbf{DSC}\\
		\hline
        \multirow{2}{*}{SwinUNETR~\cite{swin}}
        &0.9k &\XSolidBrush &67.9\\
        &11k &\XSolidBrush &68.5\\
        \hline
        \multirow{2}{*}{DiffTumor~\cite{Difftumor}}
        &0.7k &\CheckmarkBold &70.2\\
        &11k$\dagger$ &\CheckmarkBold &73.3\\
        \hline
        \multirow{5}{*}{FreeTumor}
        &0.7k &\CheckmarkBold &72.5\\
        &0.9k &\CheckmarkBold &72.9\\
        &2.2k &\CheckmarkBold &74.4\\
        &6k &\CheckmarkBold &76.3\\
        &11k &\CheckmarkBold &\cellcolor{pink}76.8\\
        \toprule[1.2pt]
    
    \end{tabular}
}{
  \caption{Average DSC when scaling up data.
  \label{table_scaling}
  \vspace{-.1in}
  }
}

\ffigbox[\FBwidth]{
\vspace{-.5in}
\includegraphics[scale=0.5]{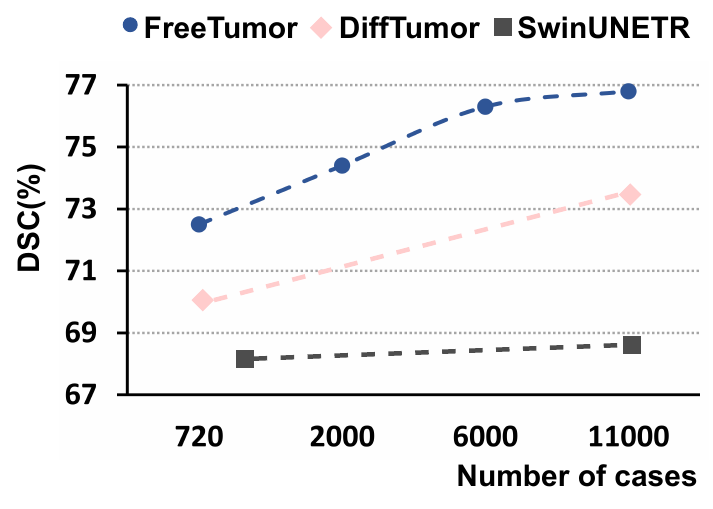}
}{
\vspace{-.2in}
  \caption{Data scaling law in tumor segmenation. The result of DiffTumor~\cite{Difftumor} on 11k is re-implement.
  \label{figure_scaling}
  \vspace{-.5in}
  }
}

\vspace{-.2in}
\end{floatrow}
\end{figure}

\textbf{Training paradigms}. As shown in Table~\ref{table_abla_training}, with only synthetic tumors, FreeTumor can also achieve competitive results. When combining real and synthetic tumors, jointly training achieves better performance. As shown in Table~\ref{table_abla_generalist}, with limited data, the specialist models used in~\cite{Syntumor,Difftumor}, \emph{i.e.}, training different models for different types of tumors, can achieve better results. However, it requires several times of training thus raising the burden. With large-scale data, the universal model gains competitive results, while it only requires one model for different types of tumors.

\textbf{Data scaling law in tumor segmentation}. As shown in Table~\ref{table_scaling} and Fig.~\ref{figure_scaling}, when scaling up the dataset, the DSC \textbf{improves marginally without synthetic tumors} (results of SwinUNETR~\cite{swin}), which means scaling up dataset with only organ labels is not effective. The pre-trained DiffTumor~\cite{Difftumor} also gains fewer improvements with 11k data. In FreeTumor, the performance is improved with the increase of data scale. However, from 6k to 11k, the improvements become marginal. We conclude that medical images (\emph{e.g.}, CT) contain more consistent information across different cases~\cite{VoCo}, \emph{i.e.}, anatomic structures of different patients are similar and the redundancy may hinder the improvements.

\section{Conclusion}

In this paper, we present FreeTumor, a simple yet effective solution for robust tumor synthesis and segmentation, which unleashes the power of large-scale unlabeled data via tumor synthesis. Specifically, FreeTumor introduces the adversarial training strategy to leverage large-scale unlabeled data in synthesis training while filtering out low-quality synthetic tumors. By scaling up data from hundreds of cases to 11k cases, FreeTumor outperforms existing methods by a large margin. In the future, we will further scale up the data and evaluate the effectiveness of FreeTumor on more types of tumors. More limitations and future directions are discussed in the appendix.


\newpage

\renewcommand{\thesection}{A}
\section{Dataset and Implementation}
\label{sec_dataset}

\begin{table*}[htbp]
    \setlength{\abovecaptionskip}{-.1in}
	\setlength{\belowcaptionskip}{-.2in}
    \begin{floatrow}
    \capbtabbox{
        \begin{tabular}{ccccc}
		\toprule[1.2pt]
		\textbf{Dataset} &\textbf{Class} &\textbf{Num.}\\
		\hline
        \textbf{\emph{With Tumor labels}}\\
        LiTS~\cite{lits} &Liv., Liv. tumor &131\\
        MSD-Pancreas~\cite{MSD} &Panc., Panc. tumor &280\\
        KiTS~\cite{kits} &Kid., Kid. tumor &489\\
        \hline
        \textbf{\emph{Without Tumor labels}}\\
        CHAOS~\cite{chaos} &Liv. &20\\
        Pancreas-CT~\cite{panc_ct} &Panc. &80\\
        BTCV~\cite{btcv} &Liv., Panc., Kid. &30\\
        Amos22~\cite{amos} &Liv., Panc., Kid. &300\\
        WORD~\cite{word} &Liv., Panc., Kid. &120\\
        Flare22~\cite{FLARE22} &Liv., Panc., Kid. &50\\
        Abdomenct-1k~\cite{abdomenct1k} &Liv., Panc., Kid. &361\\
        Flare23~\cite{FLARE22} &Liv., Panc., Kid. &4000\\
        AbdomenAtlas1.0~\cite{abdomenatlas} &Liv., Panc., Kid. &5195\\
        \hline
        Total & &11056\\
        \toprule[1.2pt]
	\end{tabular}
    }{
     \caption{The details of datasets used in FreeTumor.
     }
     \label{table_dataset}
    }
    
    \capbtabbox{
    \begin{tabular}{c|c}
    \toprule[1.2pt]
		\textbf{Scale} &\textbf{Dataset}\\
		\hline
        \textbf{720}
        &Same as DiffTumor~\cite{Difftumor}\\
        \hline
        \multirow{3}{*}{\textbf{0.9k}}
        &LiTS~\cite{lits}\\
        &MSD-Pancreas~\cite{MSD}\\
        &KiTS~\cite{kits}\\
        \hline
        \multirow{7}{*}{\textbf{2.2k}}
        &CHAOS~\cite{chaos}\\
        &Pancreas-CT~\cite{panc_ct}\\
        &BTCV~\cite{btcv}\\
        &Amos22~\cite{amos}\\
        &WORD~\cite{word}\\
        &Flare22~\cite{FLARE22}\\
        &Abdomenct-1k~\cite{abdomenct1k}\\
        \hline
        \textbf{6k} &Flare23~\cite{FLARE22}\\
        \hline
        \textbf{11k} &AbdomenAtlas1.0~\cite{abdomenatlas}\\
        \toprule[1.2pt]
    \end{tabular}
    }{
     \caption{The data scales are accumulated gradually.
     }
     \label{table_datascale}
     \small
    }
    
    \end{floatrow}
\end{table*}


The details of datasets used in FreeTumor are shown in Tables~\ref{table_dataset} and ~\ref{table_datascale}. The real tumor datasets are collected from LiTS~\cite{lits}, MSD-Pancreas~\cite{MSD}, and KiTs~\cite{kits}, which contain liver, pancreas, and kidney tumors, respectively. These three datasets are also used for validation. Thanks to the previous efforts, we can collect large-scale CT datasets with organ labels, \emph{i.e.}, CHAOS~\cite{chaos}, Pancreas-CT~\cite{panc_ct}, BTCV~\cite{btcv}, Amos22~\cite{amos}, WORD~\cite{word}, Flare22~\cite{FLARE22}, Abdomenct-1k~\cite{abdomenct1k}, AbdomenAtlas~\cite{abdomenatlas}, and Flare23~\cite{FLARE22}. We adopt consistent experiment settings with DiffTumor~\cite{Difftumor} for fair comparisons. 

We conduct experiments with data scales of \textbf{720, 0.9k, 2.2k, 6k,} and \textbf{11k}. Specifically, 720 cases are used in DiffTumor, which is a cherry-pick dataset from LiTS~\cite{lits}, MSD-Pancreas~\cite{MSD}, KiTs~\cite{kits}, CHAOS~\cite{chaos}, and Pancreas-CT~\cite{panc_ct}, where the cases without tumors are discarded. For fair comparisons, we also involve it in our experiments. Then we scaled up the data with the datasets with only organ labels. Since the organ labels are already available in these datasets, we can easily build a dataset for tumor synthesis as described in the main paper. Concretely, since we only study liver, pancreas, and kidney tumors currently, we only keep the liver, pancreas, and kidney labels in these datasets for segmentation training.

The datasets without tumor labels may also contain tumors (abnormal datasets, \emph{e.g.}, Flare23~\cite{FLARE22}), in this case, we ignore the background when calculating loss~\cite{Difftumor}. Specifically, the Flare23 dataset~\cite{FLARE22} contains only partial labels, thus we first train a segmentation model to generate pseudo labels for it. 

\textbf{Training settings}. We use Adam~\cite{adamw} optimizer with a learning rate of $1e-4$ to train the generator and classification discriminator in FreeTumor. We set 100 epochs for synthesis training. For segmentation training, we follow the settings of previous works~\cite{Difftumor,VoCo}, including \textbf{data splits and pre-processing}. Specifically, SwinUNETR~\cite{swin} is used as the backbone, and our implementation is mainly based on the open-source platform Monai~\footnote{\href{https://monai.io/}{https://monai.io/}} and Pytorch~\cite{pytorch}. All the experiments are done with an NVIDIA H800 80GB GPU.

\textbf{Safeguards}. All the datasets are licensed and public without safety concerns. The patients information are all anonymous.

\begin{figure}
	\centering
	\includegraphics[width=1\linewidth]{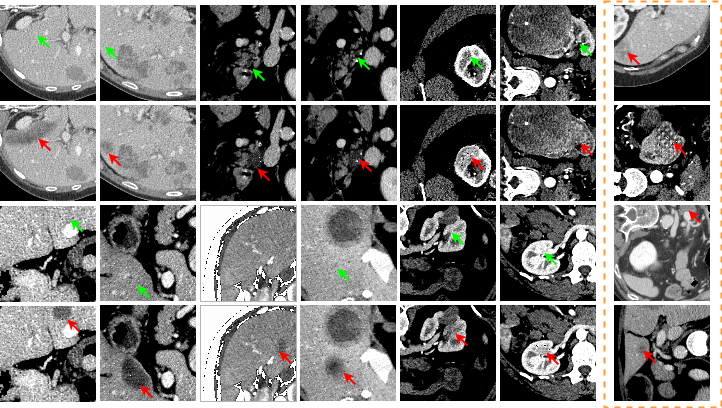}
	\caption{The \textcolor{green}{green} and \textcolor{red}{red} arrows point to the original healthy regions and the synthetic tumors, respectively. We further present some failure cases in the \textcolor{orange}{orange} box.
 \vspace{-.1in}
 }
	\label{fig_synthesis}
\vspace{-.1in}
\end{figure}

\begin{figure}
	\centering
	\includegraphics[width=1\linewidth]{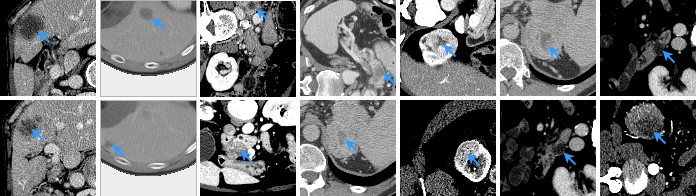}
	\caption{\textbf{\emph{Can you distinguish the real and synthetic tumors?}} The \textcolor{cyan}{blue} arrows point to the tumors, half of them are real and others are synthetic. 
 \vspace{-.1in}
 }
	\label{fig_turing}
\vspace{-.1in}
\end{figure}

\renewcommand{\thesection}{B}
\section{Tumor Synthesis: FreeTumor vs Diffusion model}

\textbf{Tumor synthesis visualization}. As shown in Fig.~\ref{fig_synthesis}, we present the synthesis results of FreeTumor. Some failure cases are also presented in the \textcolor{orange}{orange} box. \textbf{Note that} the failure cases are selected by the discriminator in FreeTumor, which are discarded in the tumor segmentation training. Although the synthesis and filtering results cannot be perfect, the negative impact of synthetic tumors has been significantly alleviated. In Fig.~\ref{fig_turing}, we present the real and synthetic tumors together. It can be seen that it is difficult to distinguish the real and synthetic tumors.

\textbf{\emph{FreeTumor vs Diffusion model}}. Diffusion models have attracted more attention in recent advanced image synthesis, which can generate high-quality, high-resolution, diversified, and realistic natural images~\cite{ldm,ddpm,vqgan}. However, in tumor synthesis, the tumors are closer to low-level abnormal noise~\cite{Syntumor, Difftumor, pixel2cancer}, which requires no complex high-level semantic control as in natural images. Thus, in the experiments, we observe that the tumor synthesis model did not require complex pipelines and techniques to train. We emphasize that the traditional GAN-based generative models can also fulfill the synthesis of tumors.

\begin{figure}
	\centering
	\includegraphics[width=1\linewidth]{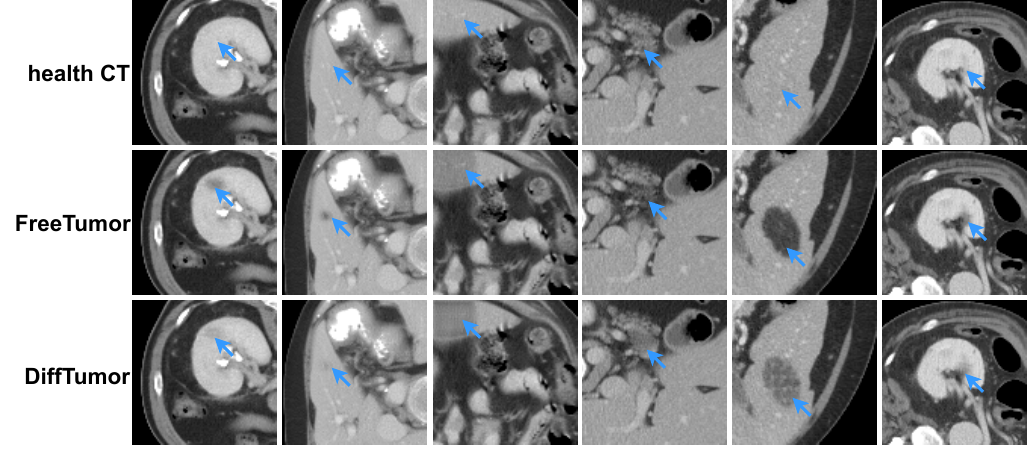}
	\caption{\textbf{\emph{FreeTumor vs DiffTumor}}~\cite{Difftumor}. \textbf{Note that} the shown data are unseen in the synthesis training of DiffTumor. We directly use the trained DiffTumor~\cite{Difftumor} model and synthesize the tumors at the same positions. The \textcolor{cyan}{blue} arrows point to the tumors. \textbf{However}, note that better visualization results do not always lead to better segmentation training~\cite{freemask}, sometimes slightly worse synthetic results can be seen as data perturbation thus may make the model more robust. We will explore it in the extension.
 \vspace{-.1in}
 }
	\label{fig_vsdiff}
\vspace{-.1in}
\end{figure}

The training of conditioned diffusion models in tumor synthesis requires adequate annotated tumor cases, which is a disadvantage that may hinder the performance of synthesis training. Specifically, the previous SOTA DiffTumor collected 10k CT to train the autoencoder~\cite{vqgan} in diffusion models, but during synthesis training (Stage 2), only 0.3k of them can be used. In the experiments, we found that DiffTumor~\cite{Difftumor} can generate excellent fake tumors in the labeled dataset(~0.3k). However, when we adapt it to \textbf{\emph{10k unseen datasets from other resources}}, the synthesis results will decrease significantly. Since medical images can be acquired from different imaging devices, a tumor synthesis generative model trained on a limited scale of datasets may fail to generalize to unseen datasets. The visualization results are shown in Fig.~\ref{fig_vsdiff}. Although scratch from the visualization results, it is still difficult to observe obvious visual differences, the segmentation results of these synthetic tumors can still reveal the performances of synthesis (discussed in the main paper). Thus, we highlight that by training with larger-scale data, \textbf{FreeTumor can acquire better synthesis results than the conditioned diffusion model}~\cite{Difftumor}.

\textbf{Remark}. We observe that better synthetic tumors (in visualization) do not always lead to better segmentation results (well segmented by the baseline segmentor). This is because the tumor segmentation is decided by not only the synthesis reality but also other random factors, \emph{e.g.}, positions, shapes, and sizes, which are brought by the random tumor mask $M$~\cite{Syntumor,Difftumor}. Thus, it is vital to develop a filtering strategy to automatically undermine the negative impacts of these random factors.

In the main paper, we have further highlighted the importance of filtering out unsatisfactory synthetic tumors, which is the main difference between FreeTumor and previous methods~\cite{Syntumor,Difftumor,pixel2cancer}. The effectiveness has been underlined in the main paper. Based on these two main advantages, \emph{i.e.}, \textbf{large-scale synthesis training and filtering strategy}, we would like to emphasize the novelty of our proposed FreeTumor compared with previous works.

\begin{figure}
	\centering
	\includegraphics[width=1\linewidth]{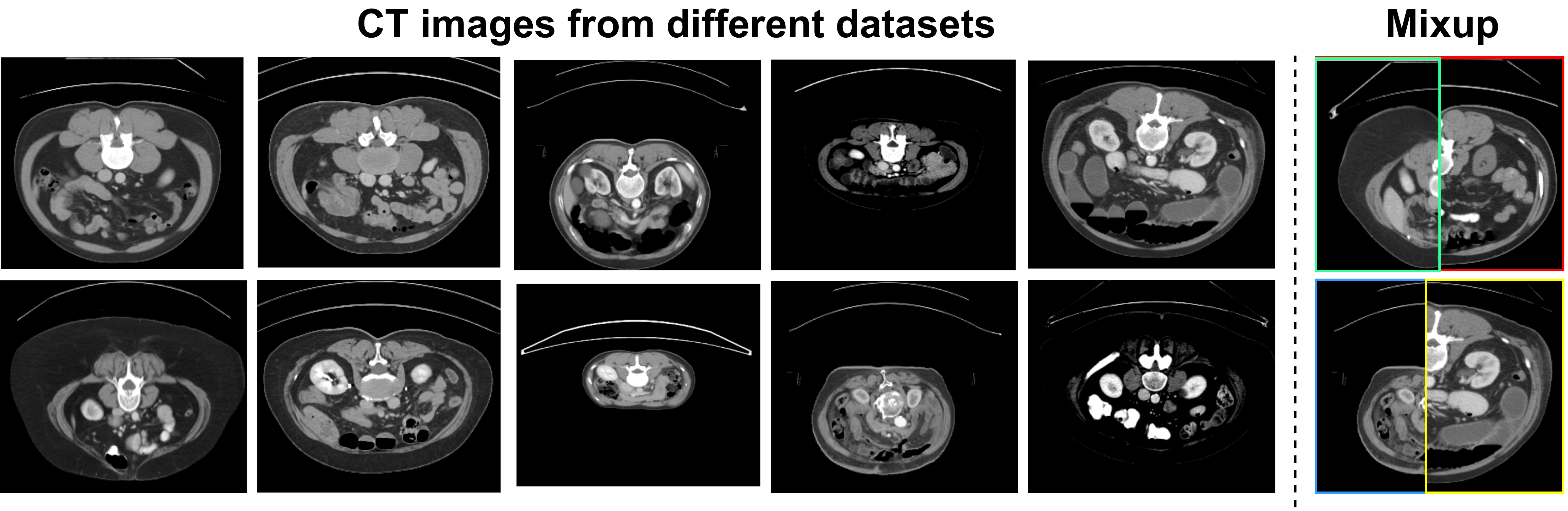}
	\caption{\textbf{Mixup for different datasets}. We observe that different datasets (from Table~\ref{table_dataset}) may contain very different charismatics, \emph{e.g}, spacing, intensity, and resolution. We use Mixup~\cite{cutmix} as strong data augmentation to bridge the gap during training.   
 \vspace{-.1in}
 }
	\label{fig_mixup}
\vspace{-.1in}
\end{figure}

\begin{figure}
	\centering
	\includegraphics[width=1\linewidth]{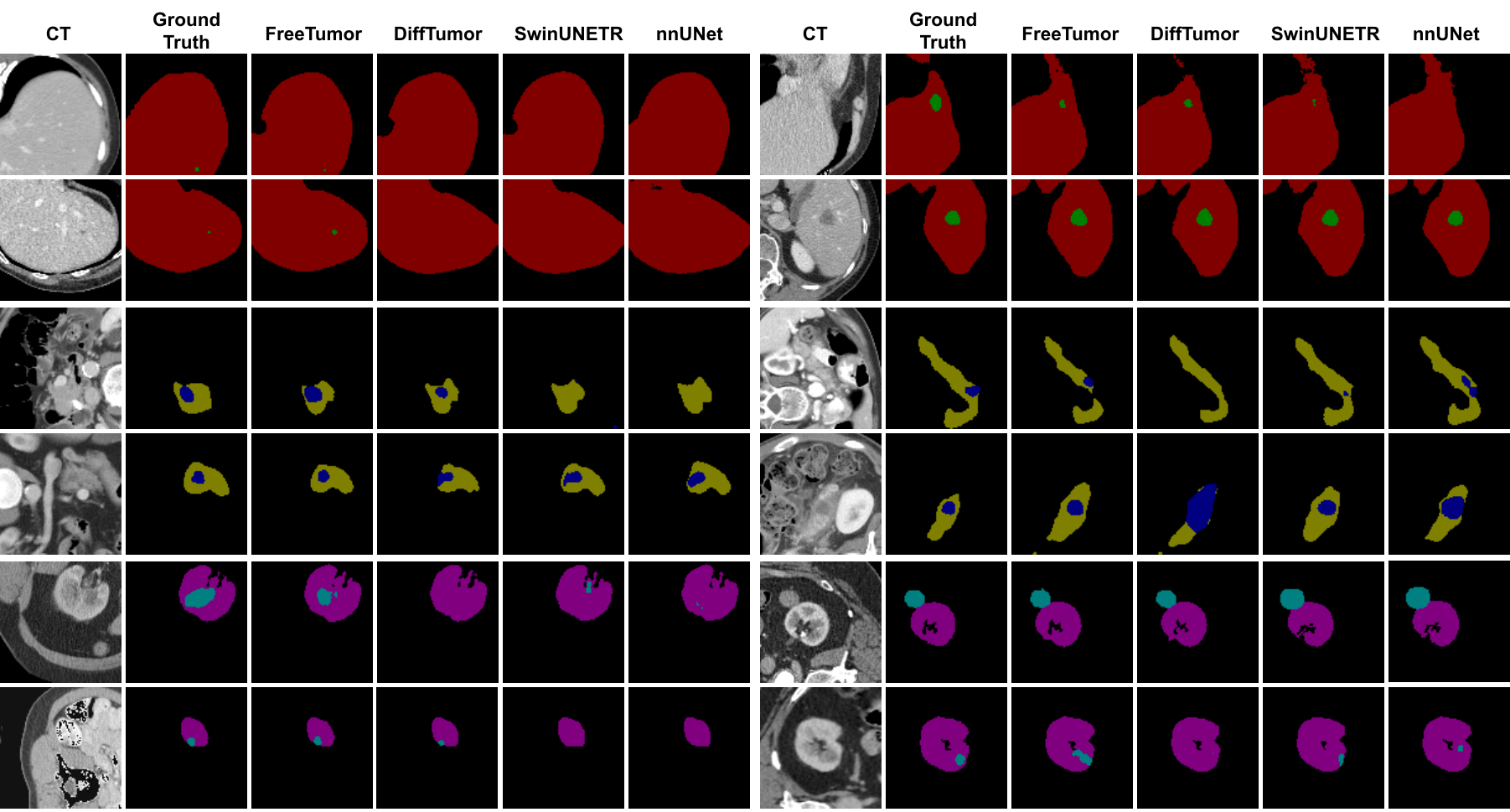}
	\caption{\textbf{Qualitative visualization results of tumor segmentation}. We further provide the segmentation results of nnUNet~\cite{nnunet}, SwinUNETR~\cite{swin}, and DiffTumor~\cite{Difftumor} for comparison. Different types of tumors are compared. Specifically, from the first row to the final row, liver, pancreas, and kidney tumors results are presented.
 \vspace{-.1in}
 }
	\label{fig_segmentation}
\vspace{-.1in}
\end{figure}

\renewcommand{\thesection}{C}
\section{Mixup for Segmentation Training}

In the main paper, we highlight the effectiveness of Mixup~\cite{cutmix} for segmentation training. As shown in Fig.~\ref{fig_mixup}, we present the visualization results of Mixup. It can be seen that in our collected 11k dataset, the images are from different resources thus appealing very different from each other, \emph{e.g.}, intensity, spacing, and resolution. To this end, motivated by previous works~\cite{CISC_R,freemask,depthanything}, we propose to use Mixup to bridge the gap between different datasets. During training, given a mixed image, patches from different datasets are optimized collaboratively. In the experiments (refer to the main paper), we found that with large-scale data, Mixup can significantly improve the segmentation performances.

\renewcommand{\thesection}{D}
\section{Tumor Segmentation Visualization}

We further provide qualitative visualizations of tumor segmentation in Fig.~\ref{fig_segmentation}. We further provide the segmentation results of nnUNet~\cite{nnunet}, SwinUNETR~\cite{swin}, and DiffTumor~\cite{Difftumor} for comparison. Specifically, visualizations of different types of tumors are presented, \emph{e.g.}, liver, pancreas, and kidney. It can be seen that FreeTumor yields a substantial improvement in segmentation performance. More importantly, tiny tumors can also be well detected, which is significant for early tumor detection. Although the results cannot be perfect, it can be seen that our proposed FreeTumor stands out from other methods with better visualization results.

\renewcommand{\thesection}{E}
\section{Public LeaderBoard Results of FLARE23}



To verify the effectiveness of FreeTumor, we further submitted results on the FLARE23 public online leaderboard\footnote{Due to the anonymous policy, we will provide the submission link upon acceptance.}, as shown in Table~\ref{table_leaderboard}. For FLARE23, there are two evaluation phases, development and final, which contain 100 and 400 cases for validation, respectively. We evaluate both of them. Specifically, FLARE23 is developed for organ and tumor segmentation. Since this paper is for tumor segmentation, we only train the tumor segmentation model and ensemble the results with a baseline organ segmentation model~\cite{swin}. In addition, since FLARE23 contains only partial labels, we use the officially provided pseudo labels for training. The tumor types in FLARE23 include liver, pancreas, kidney, stomach, and colon tumors. Since we only generate liver, pancreas, and kidney tumors for training, we guess that our improvements are mainly on these three types (although the details cannot be known). It can be seen in Table~\ref{table_leaderboard} that our FreeTumor can gain significant improvements upon previous methods. \textbf{Note that} the online results are always updating.

\begin{table}
	\setlength{\abovecaptionskip}{0.pt}
	\setlength{\belowcaptionskip}{-0.em}
	\centering
	\footnotesize
 \caption{Performances on the FLARE23 public leaderboard. We report the tumor segmentation DSC(\%). Syn. denotes whether to synthesize tumors for training.}
 \vspace{-.1in}
\begin{threeparttable}
	\begin{tabular}{cccc}
		\toprule[1.2pt]
		\multirow{2}{*}{\textbf{Method}} &\multirow{2}{*}{\textbf{Syn.}} &\multicolumn{2}{c}{\textbf{FLARE23}}\\
        \cline{3-4}
        & &Devel. &Final\\
		\hline
        nnUNet~\cite{nnunet} &\XSolidBrush &38.3 &52.9\\
        SwinUNETR~\cite{swin} &\XSolidBrush &26.7 &44.3\\
        \hline
        DiffTumor~\cite{Difftumor} &\CheckmarkBold &32.5 &41.1\\
        \rowcolor{pink}
        FreeTumor &\CheckmarkBold &\textbf{48.3} &\textbf{61.5}\\
        \toprule[1.2pt]
	\end{tabular}
    \end{threeparttable}        
\label{table_leaderboard}
\vspace{-.1in}
\end{table}%

\renewcommand{\thesection}{F}
\section{More Experiments}

\begin{table}
	\setlength{\abovecaptionskip}{0.pt}
	\setlength{\belowcaptionskip}{-0.em}
	\centering
	\footnotesize
 \caption{Evaluation of different network architectures. We report the average and variances of 5-fold data splits. * represent that the results are drawn from the original DiffTumor paper~\cite{Difftumor} appendix C.}
 \vspace{-.1in}
\begin{threeparttable}
	\begin{tabular}{ccccc}
		\toprule[1.2pt]
        \textbf{Method} &\textbf{Network} &\textbf{Liver}  &\textbf{Pancreas} &\textbf{Kidney}\\
		\hline
		\multirow{3}{*}{\textbf{Real Tumor}} 
        &UNet*~\cite{UNET} &62.5$\pm$13.6 &51.2$\pm$13.7 &72.0$\pm$8.6 \\
        &UNETR~\cite{unetr} &64.0$\pm$13.3 &55.6$\pm$12.4 &78.8$\pm$7.0\\
        &Swin-UNETR~\cite{swin} &64.7$\pm$12.0 &59.0$\pm$11.5 &80.3$\pm$6.3 \\
        \hline
		\multirow{2}{*}{\textbf{DiffTumor}*~\cite{Difftumor}} 
        &UNet~\cite{UNET} &66.5$\pm$12.9 &60.0$\pm$12.7 &79.0$\pm$7.7\\
        &Swin-UNETR~\cite{swin} &67.9$\pm$10.2 &61.0$\pm$11.6 &81.8$\pm$7.2\\
        \hline
		\multirow{3}{*}{\textbf{FreeTumor}} 
        &UNet~\cite{UNET} &69.1$\pm$5.2 &63.4$\pm$8.9 &84.1$\pm$4.0\\
        &UNETR~\cite{unetr} &73.6$\pm$4.8 &67.7$\pm$8.1 &87.6$\pm$2.0\\
        &Swin-UNETR~\cite{swin} &74.5$\pm$3.1 &68.6$\pm$7.5 &87.3$\pm$2.2\\
        \toprule[1.2pt]
	\end{tabular}
    \end{threeparttable}        
\label{table_variance}
\vspace{-.1in}
\end{table}%

\textbf{Evaluation of different network architectures}. We further evaluate the effectiveness of FreeTumor with different network architectures, \emph{i.e.}, 3D-UNet~\cite{UNET}, UNETR~\cite{unetr}, and Swin-UNETR. We also report the error bars (average and variance) of 5-fold splits. The data splits are the same as DiffTumor~\cite{Difftumor} for fair comparisons. The details are shown in Table~\ref{table_variance}. Since the tumor cases are rare in the real labeled dataset, the results of different data splits vary from each other. It can be seen that with more data for training, the variances of FreeTumor under different data splits are much smaller than that of the previous methods, which proves that our proposed method is more stable.

\textbf{Settings of ${\lambda}_{cls}$ for the balance of loss functions}. It can be seen in Table~\ref{table_lambda} that the value of ${\lambda}_{cls}$ does not make significant differences. Thus, we empirically set it as $0.1$ by balacing the scale of loss ${L}_{cls}$ and ${L}_{seg}$. 

\textbf{Different filtering strategies}. We further evaluate different paradigms of filtering strategy. In the main paper, we describe the way of using a threshold to split high- and low-quality synthetic tumors. We have further tried to use an adaptive way~\cite{freemask}, \emph{i.e.}, use the proportion $P$ as the weights of segmentation loss functions, as follows:
\begin{equation}\label{eqn_weight_loss}
    L_{seg} = \frac{1}{|N|} \sum_{i=1}^{N} P_{i} * L_{seg}(i),
\end{equation}
where $L_{seg}$ represents the segmentation loss function (a typical Dice-CE loss is used as previous works~\cite{Syntumor,Difftumor}), $P_{i}$ and $L_{seg}(i)$ denote the proportion and loss of the ${i}_{th}$ synthetic tumor, respectively. The results are shown in Table~\ref{table_adapt}. However, we found that the adaptive way does not perform better. We conclude that although with smaller weights, the low-quality tumors still contribute negatively to the segmentation training. Thus, it is more practical to dump the tumors of low proportion $P$ in segmentation training.

\begin{table*}
    \setlength{\abovecaptionskip}{-.1in}
	\setlength{\belowcaptionskip}{-.2in}
    \begin{floatrow}
    \capbtabbox{
        \begin{tabular}{cc}
		\toprule[1.2pt]
        \bm{${\lambda}_{cls}$} &\textbf{Average DSC}\\
		\hline
		1.0 &76.5\\
        0.5 &76.6\\
        0.2 &76.8\\
        0.1 &76.8\\
        \toprule[1.2pt]
	\end{tabular}
    }{
     \caption{Evaluation of ${\lambda}_{cls}$. 
     \vspace{-.1in}}
     \label{table_lambda}
    }
    
    \capbtabbox{
    \begin{tabular}{ccccc}
    \toprule[1.2pt]
		\textbf{Method} &\textbf{Synthetic} &\textbf{Filtering} &\textbf{Average DSC}\\
        \hline
        SwinUNETR~\cite{swin} &\XSolidBrush &\XSolidBrush &67.9\\
        \hline
        FreeTumor &\CheckmarkBold &\textbf{Adaptive (Eq.~\ref{eqn_weight_loss})} &73.3\\
        \hline
        FreeTumor &\CheckmarkBold &\XSolidBrush &72.3\\
        FreeTumor &\CheckmarkBold &\bm{$T=0.5$} &73.5\\
        FreeTumor &\CheckmarkBold &\bm{$T=0.7$} &\cellcolor{pink}74.4\\
        FreeTumor &\CheckmarkBold &\bm{$T=0.9$} &74.0\\
        \toprule[1.2pt]
    \end{tabular}
    }{
     \caption{Evaluation of different filtering paradigms.
     \vspace{-.1in}}
     \label{table_adapt}
     \small
    }
    \end{floatrow}
\vspace{-.1in}
\end{table*}

\textbf{Online vs Offline tumor synthesis}. As stated in the main paper, we use online tumor synthesis to generate diversified tumors. We further compare it with the offline way, \emph{i.e.}, generate tumors and corresponding labels offline, then store them for segmentation training. Thus, in each iteration, the training data are fixed. 

As in Table~\ref{table_online_offline}, we find that online generation is better. However, if we \emph{\textbf{iteratively synthesize and filter the tumors for each case until diversified and high-quality tumors are synthesized}}, the results can be competitive. Here, the measurement of high quality also refers to the filtering strategy stated in the main paper. The advantage of the offline way is that we can discard the online generation process during segmentation training. \textbf{We will release our 11k synthetic tumor dataset to facilitate the research of tumor segmentation}.

\begin{table}[htbp]
	\setlength{\abovecaptionskip}{0.pt}
	\setlength{\belowcaptionskip}{-0.em}
	\centering
	\footnotesize
 \caption{Online vs Offline tumor synthesis. `Iterative' denotes conducting synthesis and filtering for each case iteratively.}
 \vspace{-.1in}
\begin{threeparttable}
	\begin{tabular}{cccc}
		\toprule[1.2pt]
        \textbf{Data Scale} &\textbf{Manner} &\textbf{Iterative} &\textbf{Average DSC}\\
		\hline
    2.2k &online & &74.4\\
    2.2k &offline & &72.0\\
    2.2k &offline &\CheckmarkBold &73.7\\
    \hline
    11k &online & &76.8\\
    11k &offline & &73.3\\
    11k &offline &\CheckmarkBold &76.6\\
        \toprule[1.2pt]
	\end{tabular}
    \end{threeparttable}        
\label{table_online_offline}
\vspace{-.2in}
\end{table}

\begin{table}[htbp]
	\setlength{\abovecaptionskip}{0.pt}
	\setlength{\belowcaptionskip}{-0.em}
	\centering
	\footnotesize
 \caption{Evaluation of the zero-shot tumor synthesis ability. We report the average DSC. We directly use the trained DiffTumor~\cite{Difftumor} models to test the ability of DiffTumor~\cite{Difftumor}. For efficiency, we only test the performance of SwinUnetr~\cite{swinunetr} and use the same data split.}
 \vspace{-.1in}
\begin{threeparttable}
	\begin{tabular}{c|cc|cc|cc}
		\toprule[1.2pt]
        \textbf{Method} &\textbf{Liv-->Panc} &\textbf{Liv-->Kid}  &\textbf{Panc-->Liv} &\textbf{Panc-->Kid} &\textbf{Kid-->Liv} &\textbf{Kid-->Panc}\\
		\hline
		SynTumor~\cite{Syntumor}
        &42.5 &67.9 &- &- &- &-\\
		DiffTumor~\cite{Difftumor}
        &55.4 &71.2 &61.5 &71.3 &63.8 &49.8\\
        \rowcolor{pink}
		FreeTumor
        &62.3 &82.6 &68.1 &81.2 &70.4 &59.7\\
        \toprule[1.2pt]
	\end{tabular}
    \end{threeparttable}        
\label{table_zeroshot}
\vspace{-.1in}
\end{table}%

\textbf{Zero-shot tumor synthesis ability}. We further evaluate the zero-shot ability of FreeTumor, \emph{i.e.}, synthesis training on one specific type of tumor and directly adapt the model to synthesize other types of tumors. The results are presented in Table~\ref{table_zeroshot}. It can be seen that although the results will drop when adapting to other types, FreeTumor still surpasses SynTumor~\cite{Syntumor} and DiffTumor~\cite{Difftumor} by a large margin. This is because the filtering strategy will help to control the synthesis results in FreeTumor and alleviate the negative impact of low-quality synthetic results. In addition, our FreeTumor synthesizes tumors by estimating the value distances between organs and tumors, thus the tumor synthesis is also robust to the corresponding organs. While DiffTumor~\cite{Difftumor} generates tumors by directly reconstructing the values of tumors, thus may fail to generalize from one type to another.

\renewcommand{\thesection}{G}
\section{Limitations and Future Directions}

Although promising results are demonstrated, there are still several limitations of FreeTumor that can be further extended in the future: 
\begin{itemize}

\item In the current version, we only study the three most common types of tumors, \emph{i.e.}, liver, pancreas, and kidney. In the extension, we will further collect datasets for studying stomach, colon, brain, and lung cancer, making a further step towards better tumor segmentation models. 

\item Instead of pursuing new synthesis techniques, we explore the tumor synthesis paradigm by highlighting the power of large-scale data. In the extension, we will dive deeper into the tumor characteristics and explore more advanced tumor synthesis techniques to benefit the following tumor segmentation training.

\item It is tedious for radiologists to evaluate the quality of tumor synthesis in a large-scale dataset. In the future, we will further explore a more efficient way to verify the quality with the help of some experienced radiologists. 

\item In the experiments, we observe that when involving large-scale datasets from different sources, the tumor segmentation training becomes unstable. In the extension, we will further explore the data-processing techniques and training paradigms to facilitate the tumor segmentation training of large-scale data.

\item When scaling-up data from 6k to 11k, the improvements become marginal. In the extension, we will further study the training paradigms to effectively leverage large-scale data and explore the bound of data scaling law in tumor segmentation.

\end{itemize}


{\small
\bibliographystyle{ieee_fullname}
\bibliography{references}

\begin{thebibliography}{10}\itemsep=-1pt

\bibitem{MSD}
Michela Antonelli et~al.
\newblock The medical segmentation decathlon.
\newblock {\em Nature Commun.}, 13(1):4128, 2022.

\bibitem{assemlal2011recent}
Haz-Edine Assemlal et~al.
\newblock Recent advances in diffusion mri modeling: Angular and radial reconstruction.
\newblock {\em Medical Image Analy.}, 15(4):369--396, 2011.

\bibitem{baradad2021learning}
Manel Baradad~Jurjo et~al.
\newblock Learning to see by looking at noise.
\newblock {\em NIPS}, 34:2556--2569, 2021.

\bibitem{lits}
Patrick Bilic et~al.
\newblock The liver tumor segmentation benchmark (lits).
\newblock {\em Medical Image Analy.}, 84:102680, 2023.

\bibitem{billot2023synthseg}
Benjamin Billot et~al.
\newblock Synthseg: Segmentation of brain mri scans of any contrast and resolution without retraining.
\newblock {\em Medical Image Analy.}, 86:102789, 2023.

\bibitem{chen2019synergistic}
Cheng Chen et~al.
\newblock Synergistic image and feature adaptation: Towards cross-modality domain adaptation for medical image segmentation.
\newblock In {\em AAAI}, pages 865--872, 2019.

\bibitem{chen2020unsupervised}
Cheng Chen et~al.
\newblock Unsupervised bidirectional cross-modality adaptation via deeply synergistic image and feature alignment for medical image segmentation.
\newblock {\em IEEE Trans. Medical Imag.}, 39(7):2494--2505, 2020.

\bibitem{Difftumor}
Qi Chen, Xiaoxi Chen, Haorui Song, Zhiwei Xiong, Alan Yuille, Chen Wei, and Zongwei Zhou.
\newblock Towards generalizable tumor synthesis.
\newblock In {\em CVPR}, 2024.

\bibitem{chung2022mr}
Hyungjin Chung, Eun~Sun Lee, and Jong~Chul Ye.
\newblock Mr image denoising and super-resolution using regularized reverse diffusion.
\newblock {\em IEEE Trans. Medical Imag.}, 42(4):922--934, 2022.

\bibitem{dar2019image}
Salman~UH Dar et~al.
\newblock Image synthesis in multi-contrast mri with conditional generative adversarial networks.
\newblock {\em IEEE Trans. Medical Imag.}, 38(10):2375--2388, 2019.

\bibitem{du2023boosting}
Shiyi Du et~al.
\newblock Boosting dermatoscopic lesion segmentation via diffusion models with visual and textual prompts.
\newblock {\em arXiv preprint arXiv:2310.02906}, 2023.

\bibitem{vqgan}
Patrick Esser, Robin Rombach, and Bjorn Ommer.
\newblock Taming transformers for high-resolution image synthesis.
\newblock In {\em CVPR}, pages 12873--12883, 2021.

\bibitem{geng2021content}
Mufeng Geng et~al.
\newblock Content-noise complementary learning for medical image denoising.
\newblock {\em IEEE Trans. Medical Imag.}, 41(2):407--419, 2021.

\bibitem{copypaste}
Golnaz Ghiasi et~al.
\newblock Simple copy-paste is a strong data augmentation method for instance segmentation.
\newblock In {\em CVPR}, pages 2918--2928, 2021.

\bibitem{GAN}
Ian Goodfellow et~al.
\newblock Generative adversarial nets.
\newblock {\em NIPS}, 27, 2014.

\bibitem{han2019synthesizing}
Changhee Han et~al.
\newblock Synthesizing diverse lung nodules wherever massively: 3d multi-conditional gan-based ct image augmentation for object detection.
\newblock In {\em 3DV}, pages 729--737. IEEE, 2019.

\bibitem{swinunetr}
Ali Hatamizadeh et~al.
\newblock Swin unetr: Swin transformers for semantic segmentation of brain tumors in mri images.
\newblock In {\em MICCAIW}, pages 272--284, 2021.

\bibitem{unetr}
Ali Hatamizadeh et~al.
\newblock Unetr: Transformers for 3d medical image segmentation.
\newblock In {\em WACV}, pages 574--584, 2022.

\bibitem{kits}
Nicholas Heller et~al.
\newblock The kits21 challenge: Automatic segmentation of kidneys, renal tumors, and renal cysts in corticomedullary-phase ct, 2023.

\bibitem{ho2022video}
Jonathan Ho et~al.
\newblock Video diffusion models.
\newblock {\em NIPS}, 35:8633--8646, 2022.

\bibitem{ddpm}
Jonathan Ho, Ajay Jain, and Pieter Abbeel.
\newblock Denoising diffusion probabilistic models.
\newblock {\em NIPS}, 33:6840--6851, 2020.

\bibitem{ho2022classifier}
Jonathan Ho and Tim Salimans.
\newblock Classifier-free diffusion guidance.
\newblock {\em arXiv preprint arXiv:2207.12598}, 2022.

\bibitem{Syntumor}
Qixin Hu, Yixiong Chen, Junfei Xiao, Shuwen Sun, Jieneng Chen, Alan~L Yuille, and Zongwei Zhou.
\newblock Label-free liver tumor segmentation.
\newblock In {\em CVPR}, pages 7422--7432, 2023.

\bibitem{nnunet}
Fabian Isensee, Paul~F Jaeger, Simon~AA Kohl, Jens Petersen, and Klaus~H Maier-Hein.
\newblock nnu-net: a self-configuring method for deep learning-based biomedical image segmentation.
\newblock {\em Nature Methods}, 18(2):203--211, 2021.

\bibitem{pix2pix}
Phillip Isola, Jun-Yan Zhu, Tinghui Zhou, and Alexei~A Efros.
\newblock Image-to-image translation with conditional adversarial networks.
\newblock In {\em CVPR}, pages 1125--1134, 2017.

\bibitem{amos}
Yuanfeng Ji et~al.
\newblock Amos: A large-scale abdominal multi-organ benchmark for versatile medical image segmentation.
\newblock {\em NIPS}, 35:36722--36732, 2022.

\bibitem{ZePT}
Yankai Jiang, Zhongzhen Huang, Rongzhao Zhang, Xiaofan Zhang, and Shaoting Zhang.
\newblock Zept: Zero-shot pan-tumor segmentation via query-disentangling and self-prompting.
\newblock In {\em CVPR}, 2024.

\bibitem{jin2021free}
Qiangguo Jin, Hui Cui, Changming Sun, Zhaopeng Meng, and Ran Su.
\newblock Free-form tumor synthesis in computed tomography images via richer generative adversarial network.
\newblock {\em Knowledge-Based Systems}, 218:106753, 2021.

\bibitem{kataoka2022replacing}
Hirokatsu Kataoka et~al.
\newblock Replacing labeled real-image datasets with auto-generated contours.
\newblock In {\em CVPR}, pages 21232--21241, 2022.

\bibitem{chaos}
A~Emre Kavur, N~Sinem Gezer, Bar{\i}{\c{s}}, et~al.
\newblock Chaos challenge-combined (ct-mr) healthy abdominal organ segmentation.
\newblock {\em Medical Image Analy.}, 69:101950, 2021.

\bibitem{kazerouni2023diffusion}
Amirhossein Kazerouni et~al.
\newblock Diffusion models in medical imaging: A comprehensive survey.
\newblock {\em Medical Image Analy.}, page 102846, 2023.

\bibitem{vqvae}
Diederik~P Kingma and Max Welling.
\newblock Auto-encoding variational bayes.
\newblock {\em arXiv preprint arXiv:1312.6114}, 2013.

\bibitem{pixel2cancer}
Yuxiang Lai, Xiaoxi Chen, Angtian Wang, Alan Yuille, and Zongwei Zhou.
\newblock From pixel to cancer: Cellular automata in computed tomography.
\newblock In {\em MICCAI}, 2024.

\bibitem{btcv}
Bennett Landman et~al.
\newblock Miccai multi-atlas labeling beyond the cranial vault--workshop and challenge.
\newblock In {\em MICCAIW}, volume~5, page~12, 2015.

\bibitem{li2023well}
Wenxuan Li, Alan Yuille, and Zongwei Zhou.
\newblock How well do supervised models transfer to 3d image segmentation?
\newblock In {\em ICLR}, 2024.

\bibitem{H-Dense}
Xiaomeng Li, Hao Chen, Xiaojuan Qi, Qi Dou, Chi-Wing Fu, and Pheng-Ann Heng.
\newblock H-denseunet: hybrid densely connected unet for liver and tumor segmentation from ct volumes.
\newblock {\em IEEE Trans. Medical Imag.}, 37(12):2663--2674, 2018.

\bibitem{li2023zero}
Yunxiang Li et~al.
\newblock Zero-shot medical image translation via frequency-guided diffusion models.
\newblock {\em IEEE Trans. Medical Imag.}, 2023.

\bibitem{clipdriven}
Jie Liu et~al.
\newblock Clip-driven universal model for organ segmentation and tumor detection.
\newblock In {\em ICCV}, pages 21152--21164, 2023.

\bibitem{liu2023multi}
Qiang Liu et~al.
\newblock A multi-level label-aware semi-supervised framework for remote sensing scene classification.
\newblock {\em IEEE Trans. Geosci. Remote Sens.}, 2023.

\bibitem{adamw}
Ilya Loshchilov and Frank Hutter.
\newblock Decoupled weight decay regularization.
\newblock {\em arXiv preprint arXiv:1711.05101}, 2017.

\bibitem{word}
Xiangde Luo et~al.
\newblock {WORD}: A large scale dataset, benchmark and clinical applicable study for abdominal organ segmentation from ct image.
\newblock {\em Medical Image Analy.}, 82:102642, 2022.

\bibitem{lyu2022pseudo}
Fei Lyu et~al.
\newblock Pseudo-label guided image synthesis for semi-supervised covid-19 pneumonia infection segmentation.
\newblock {\em IEEE Trans. Medical Imag.}, 42(3):797--809, 2022.

\bibitem{abdomenct1k}
Jun Ma et~al.
\newblock Abdomenct-1k: Is abdominal organ segmentation a solved problem?
\newblock {\em IEEE Trans. Pattern Analy. Mach. Intell.}, 44(10):6695--6714, 2021.

\bibitem{FLARE22}
Jun Ma et~al.
\newblock Unleashing the strengths of unlabeled data in pan-cancer abdominal organ quantification: the flare22 challenge.
\newblock {\em arXiv preprint arXiv:2308.05862}, 2023.

\bibitem{ma2021structure}
Yuhui Ma et~al.
\newblock Structure and illumination constrained gan for medical image enhancement.
\newblock {\em IEEE Trans. Medical Imag.}, 40(12):3955--3967, 2021.

\bibitem{nie2022diffusion}
Weili Nie et~al.
\newblock Diffusion models for adversarial purification.
\newblock {\em arXiv preprint arXiv:2205.07460}, 2022.

\bibitem{ozbey2023unsupervised}
Muzaffer {\"O}zbey et~al.
\newblock Unsupervised medical image translation with adversarial diffusion models.
\newblock {\em IEEE Trans. Medical Imag.}, 2023.

\bibitem{SPADE}
Taesung Park, Ming-Yu Liu, Ting-Chun Wang, and Jun-Yan Zhu.
\newblock Semantic image synthesis with spatially-adaptive normalization.
\newblock In {\em CVPR}, pages 2337--2346, 2019.

\bibitem{pytorch}
Adam Paszke et~al.
\newblock Pytorch: An imperative style, high-performance deep learning library.
\newblock {\em NIPS}, 32, 2019.

\bibitem{abdomenatlas}
Chongyu Qu et~al.
\newblock Abdomenatlas-8k: Annotating 8,000 ct volumes for multi-organ segmentation in three weeks.
\newblock {\em NIPS}, 36, 2024.

\bibitem{richter2016playing}
Stephan~R Richter et~al.
\newblock Playing for data: Ground truth from computer games.
\newblock In {\em ECCV}, pages 102--118, 2016.

\bibitem{ldm}
Robin Rombach, Andreas Blattmann, Dominik Lorenz, Patrick Esser, and Bj{\"o}rn Ommer.
\newblock High-resolution image synthesis with latent diffusion models.
\newblock In {\em CVPR}, pages 10684--10695, 2022.

\bibitem{UNET}
Olaf Ronneberger, Philipp Fischer, and Thomas Brox.
\newblock U-net: Convolutional networks for biomedical image segmentation.
\newblock In {\em MICCAI}, pages 234--241, 2015.

\bibitem{ros2016synthia}
German Ros et~al.
\newblock The synthia dataset: A large collection of synthetic images for semantic segmentation of urban scenes.
\newblock In {\em CVPR}, pages 3234--3243, 2016.

\bibitem{panc_ct}
H. Roth et~al.
\newblock Data from pancreas-ct.
\newblock {\em The Cancer Imaging Archive}, 2016.

\bibitem{sariyildiz2023fake}
Mert~B{\"u}lent Sar{\i}y{\i}ld{\i}z et~al.
\newblock Fake it till you make it: Learning transferable representations from synthetic imagenet clones.
\newblock In {\em CVPR}, pages 8011--8021, 2023.

\bibitem{adversarialdiffusion}
Axel Sauer et~al.
\newblock Adversarial diffusion distillation.
\newblock {\em arXiv preprint arXiv:2311.17042}, 2023.

\bibitem{shin2018abnormal}
Younghak Shin et~al.
\newblock Abnormal colon polyp image synthesis using conditional adversarial networks for improved detection performance.
\newblock {\em IEEE Access}, 6:56007--56017, 2018.

\bibitem{siddiquee2019learning}
Siddiquee et~al.
\newblock Learning fixed points in generative adversarial networks: From image-to-image translation to disease detection and localization.
\newblock In {\em ICCV}, pages 191--200, 2019.

\bibitem{souly2017semi}
Nasim Souly et~al.
\newblock Semi supervised semantic segmentation using generative adversarial network.
\newblock In {\em ICCV}, pages 5688--5696, 2017.

\bibitem{oasis}
Vadim Sushko, Edgar Sch{\"o}nfeld, Dan Zhang, Juergen Gall, Bernt Schiele, and Anna Khoreva.
\newblock Oasis: only adversarial supervision for semantic image synthesis.
\newblock {\em IJCV}, 130(12):2903--2923, 2022.

\bibitem{takashima2023visual}
Sora Takashima et~al.
\newblock Visual atoms: Pre-training vision transformers with sinusoidal waves.
\newblock In {\em CVPR}, pages 18579--18588, 2023.

\bibitem{swin}
Yucheng Tang et~al.
\newblock Self-supervised pre-training of swin transformers for 3d medical image analysis.
\newblock In {\em CVPR}, pages 20730--20740, 2022.

\bibitem{wang2022anomaly}
Hualin Wang et~al.
\newblock Anomaly segmentation in retinal images with poisson-blending data augmentation.
\newblock {\em Medical Image Analy.}, 81:102534, 2022.

\bibitem{wang2021development}
Meiyun Wang, Fangfang Fu, et~al.
\newblock Development of an ai system for accurately diagnose hepatocellular carcinoma from computed tomography imaging data.
\newblock {\em Briti. Jour. of Cancer}, 125(8):1111--1121, 2021.

\bibitem{wang2024do}
Yifei Wang, Jizhe Zhang, and Yisen Wang.
\newblock Do generated data always help contrastive learning?
\newblock In {\em ICLR}, 2024.

\bibitem{wolleb2022diffusion}
Julia Wolleb et~al.
\newblock Diffusion models for medical anomaly detection.
\newblock In {\em MICCAI}, pages 35--45. Springer, 2022.

\bibitem{wu2024medsegdiff}
Junde Wu et~al.
\newblock Medsegdiff-v2: Diffusion-based medical image segmentation with transformer.
\newblock In {\em AAAI}, volume~38, pages 6030--6038, 2024.

\bibitem{DBFNet}
Linshan Wu et~al.
\newblock Deep bilateral filtering network for point-supervised semantic segmentation in remote sensing images.
\newblock {\em IEEE Trans. Image Process.}, 31:7419--7434, 2022.

\bibitem{agmm++}
Linshan Wu et~al.
\newblock Modeling the label distributions for weakly-supervised semantic segmentation.
\newblock {\em arXiv preprint arXiv:2403.13225}, 2024.

\bibitem{CISC_R}
Linshan Wu, Leyuan Fang, Xingxin He, Min He, Jiayi Ma, and Zhun Zhong.
\newblock Querying labeled for unlabeled: Cross-image semantic consistency guided semi-supervised semantic segmentation.
\newblock {\em IEEE Trans. Pattern Anal. Mach. Intell.}, 45(7):8827--8844, Jul. 2023.

\bibitem{DCA}
Linshan Wu, Ming Lu, and Leyuan Fang.
\newblock Deep covariance alignment for domain adaptive remote sensing image segmentation.
\newblock {\em IEEE Trans. Geosci. Remote Sens.}, 60:1--11, 2022.

\bibitem{AGMM}
Linshan Wu, Zhun Zhong, Leyuan Fang, Xingxin He, Qiang Liu, Jiayi Ma, and Hao Chen.
\newblock Sparsely annotated semantic segmentation with adaptive gaussian mixtures.
\newblock In {\em CVPR}, pages 15454--15464, 2023.

\bibitem{VoCo}
Linshan Wu, Jiaxin Zhuang, and Hao Chen.
\newblock Voco: A simple-yet-effective volume contrastive learning framework for 3d medical image analysis.
\newblock In {\em CVPR}, 2024.

\bibitem{wyatt2022anoddpm}
Julian Wyatt, Adam Leach, Sebastian~M Schmon, and Chris~G Willcocks.
\newblock Anoddpm: Anomaly detection with denoising diffusion probabilistic models using simplex noise.
\newblock In {\em CVPR}, pages 650--656, 2022.

\bibitem{Xiang_2023_CVPR}
Tiange Xiang et~al.
\newblock Squid: Deep feature in-painting for unsupervised anomaly detection.
\newblock In {\em CVPR}, pages 23890--23901, June 2023.

\bibitem{freestyle}
Han Xue et~al.
\newblock Freestyle layout-to-image synthesis.
\newblock In {\em CVPR}, pages 14256--14266, 2023.

\bibitem{yang2022st++}
Lihe Yang et~al.
\newblock St++: Make self-training work better for semi-supervised semantic segmentation.
\newblock In {\em CVPR}, pages 4268--4277, 2022.

\bibitem{yang2023revisiting}
Lihe Yang et~al.
\newblock Revisiting weak-to-strong consistency in semi-supervised semantic segmentation.
\newblock In {\em CVPR}, pages 7236--7246, 2023.

\bibitem{yang2023shrinking}
Lihe Yang et~al.
\newblock Shrinking class space for enhanced certainty in semi-supervised learning.
\newblock In {\em ICCV}, pages 16187--16196, 2023.

\bibitem{depthanything}
Lihe Yang, Bingyi Kang, Zilong Huang, Xiaogang Xu, Jiashi Feng, and Hengshuang Zhao.
\newblock Depth anything: Unleashing the power of large-scale unlabeled data.
\newblock In {\em CVPR}, 2024.

\bibitem{freemask}
Lihe Yang, Xiaogang Xu, Bingyi Kang, Yinghuan Shi, and Hengshuang Zhao.
\newblock Freemask: Synthetic images with dense annotations make stronger segmentation models.
\newblock {\em NIPS}, 36, 2024.

\bibitem{yao2021label}
Qingsong Yao, Li Xiao, Peihang Liu, and S~Kevin Zhou.
\newblock Label-free segmentation of covid-19 lesions in lung ct.
\newblock {\em IEEE Trans. Medical Imag.}, 40(10):2808--2819, 2021.

\bibitem{cutmix}
Sangdoo Yun et~al.
\newblock Cutmix: Regularization strategy to train strong classifiers with localizable features.
\newblock In {\em ICCV}, pages 6023--6032, 2019.

\bibitem{controlnet}
Lvmin Zhang, Anyi Rao, and Maneesh Agrawala.
\newblock Adding conditional control to text-to-image diffusion models.
\newblock In {\em ICCV}, pages 3836--3847, 2023.

\bibitem{zhang2021datasetgan}
Yuxuan Zhang et~al.
\newblock Datasetgan: Efficient labeled data factory with minimal human effort.
\newblock In {\em CVPR}, pages 10145--10155, 2021.

\bibitem{zhong2020random}
Zhun Zhong et~al.
\newblock Random erasing data augmentation.
\newblock In {\em AAAI}, volume~34, pages 13001--13008, 2020.

\bibitem{zhong2022adversarial}
Zhun Zhong et~al.
\newblock Adversarial style augmentation for domain generalized urban-scene segmentation.
\newblock {\em NIPS}, 35:338--350, 2022.

\bibitem{zhou2021review}
S~Kevin Zhou, Hayit Greenspan, et~al.
\newblock A review of deep learning in medical imaging: Imaging traits, technology trends, case studies with progress highlights, and future promises.
\newblock {\em Proc. of the IEEE}, 109(5):820--838, 2021.

\bibitem{cyclegan}
Jun-Yan Zhu, Taesung Park, Phillip Isola, and Alexei~A Efros.
\newblock Unpaired image-to-image translation using cycle-consistent adversarial networks.
\newblock In {\em ICCV}, pages 2223--2232, 2017.

\bibitem{mim}
Jiaxin Zhuang et~al.
\newblock Mim: Mask in mask self-supervised pre-training for 3d medical image analysis.
\newblock {\em arXiv preprint arXiv:2404.15580}, 2024.

\bibitem{mmwhs}
Xiahai Zhuang.
\newblock Multivariate mixture model for myocardial segmentation combining multi-source images.
\newblock {\em IEEE Trans. Pattern Analy. Mach. Intell.}, 41(12):2933--2946, 2018.

\end{thebibliography}
}

\end{document}